%% file: main.tex
\documentclass[sigconf, nonacm]{acmart}
\usepackage{balance}
\newcommand\vldbdoi{XX.XX/XXX.XX}
\newcommand\vldbpages{XXX-XXX}
\newcommand\vldbvolume{14}
\newcommand\vldbissue{1}
\newcommand\vldbyear{2020}
\newcommand\vldbauthors{\authors}
\newcommand\vldbtitle{\shorttitle}
\newcommand\vldbavailabilityurl{https://github.com/xuanyang19/Local_Shapley_Supplemental_Material}
\newcommand\vldbpagestyle{plain}

\input{lib_package}

\input{lib_command}

\begin{document}
\title{Local Shapley: Model-Induced Locality and Optimal Reuse in Data Valuation}
\input{lib_author}

\input{sections_0_abstract_index}

\maketitle

\input{lib_vldb}

\input{sections_1_introduction_index}
\input{sections_2_related_index}
\input{sections_3_preliminary_index}
\input{sections_4_exact_index}

\input{sections_5_approximation_index}
\input{sections_6_experiment_index}

\input{sections_conclusion}
\balance
\bibliographystyle{ACM-Reference-Format}
\bibliography{reference}
\clearpage
\appendix
\input{sections_a_proof_main}

\input{sections_b_experiment_main}

\end{document}

%% file: lib_package.tex
\usepackage{amsmath}
\usepackage{amsfonts}
\usepackage{amsthm}
\usepackage{mathtools}
\usepackage{xspace}
\usepackage{multirow} 
\usepackage{xcolor} 
\usepackage{pifont}
\usepackage{subfigure}
\usepackage{bbm}
\usepackage{algorithm}
\usepackage[noend]{algpseudocode}
\usepackage[overload]{empheq}

\usepackage{microtype}
\usepackage{nicefrac}      
\usepackage{graphicx}         

\usepackage{hyperref}
\usepackage{url}

\usepackage{booktabs}
\usepackage{siunitx}
\usepackage{enumitem}

%% file: lib_command.tex
\theoremstyle{plain}
\newtheorem{theorem}{Theorem}
\newtheorem{proposition}{Proposition}
\newtheorem{lemma}{Lemma}

\newtheorem{definition}{Definition}
\newtheorem{corollary}{Corollary}
\newtheorem{assumption}{Assumption}
\newtheorem{remark}{Remark}

\newtheorem{innercustomthm}{Theorem}
\newenvironment{customthm}[1]
  {\renewcommand\theinnercustomthm{#1}\innercustomthm}
  {\endinnercustomthm}

\newtheorem{innercustomprop}{Proposition}
\newenvironment{customprop}[1]
  {\renewcommand\theinnercustomprop{#1}\innercustomprop}
  {\endinnercustomprop}

\newtheorem{innercustomlemma}{Lemma}
\newenvironment{customlemma}[1]
  {\renewcommand\theinnercustomlemma{#1}\innercustomlemma}
  {\endinnercustomlemma}

\newtheorem{innercustomcorollary}{Corollary}
\newenvironment{customcorollary}[1]{\renewcommand\theinnercustomcorollary{#1}\innercustomcorollary}
  {\endinnercustomcorollary}

\newtheorem*{theorem*}{Theorem}
\newtheorem*{proposition*}{Proposition}
\newtheorem*{lemma*}{Lemma}
\newtheorem*{property*}{Property}
\newtheorem*{definition*}{Definition}

\theoremstyle{definition}

\newcommand{\nop}[1]{}

\usepackage{xcolor}

{\normalfont\bfseries}

\setlist[itemize]{leftmargin=16pt}
\setlist[enumerate]{leftmargin=16pt}

%% file: lib_author.tex
\author{Xuan Yang}
\affiliation{%
  \institution{Duke University}
  \city{Durham}
  \state{NC}
  \country{USA}
}
\email{xuan.yang@duke.edu}

\author{Hsi-Wen Chen}
\affiliation{%
  \institution{National Taiwan University}
  \city{Taipei}
  \country{Taiwan}
}
\email{hwchen@arbor.ee.ntu.edu.tw}


\author{Ming-Syan Chen}
\affiliation{%
  \institution{National Taiwan University}
  \city{Taipei}
  \country{Taiwan}
}
\email{mschen@ntu.edu.tw}

\author{Jian Pei}
\affiliation{%
  \institution{Duke University}
  \city{Durham}
  \state{NC}
  \country{USA}
}
\email{j.pei@duke.edu}

%% file: sections_0_abstract_index.tex
\begin{abstract} 
The Shapley value provides a principled foundation for data valuation, but exact computation is \#P-hard due to the exponential coalition space. Existing accelerations remain global and ignore a structural property of modern predictors: for a given test instance, only a small subset of training points influences the prediction. We formalize this \textbf{model-induced locality} through support sets defined by the model’s computational pathway (e.g., neighbors in KNN, leaves in trees, receptive fields in GNNs), showing that Shapley computation can be projected onto these supports without loss when locality is exact. This reframes Shapley evaluation as a structured data processing problem over overlapping support-induced subset families rather than exhaustive coalition enumeration. We prove that the intrinsic complexity of Local Shapley is governed by the number of \emph{distinct influential subsets}, establishing an information-theoretic lower bound on retraining operations. Guided by this result, we propose \textbf{LSMR} (\textbf{L}ocal \textbf{S}hapley via \textbf{M}odel \textbf{R}euse), an optimal subset-centric algorithm that trains each influential subset exactly once via support mapping and pivot scheduling. For larger supports, we develop \textbf{LSMR-A}, a reuse-aware Monte Carlo estimator that remains unbiased with exponential concentration, with runtime determined by the number of distinct sampled subsets rather than total draws. Experiments across multiple model families demonstrate substantial retraining reductions and speedups while preserving high valuation fidelity.

\end{abstract}

%% file: lib_vldb.tex
\pagestyle{\vldbpagestyle}
\begingroup\small\noindent\raggedright\textbf{PVLDB Reference Format:}\\
\vldbauthors. \vldbtitle. PVLDB, \vldbvolume(\vldbissue): \vldbpages, \vldbyear.\\
\href{https://doi.org/\vldbdoi}{doi:\vldbdoi}
\endgroup
\begingroup
\renewcommand\thefootnote{}\footnote{\noindent
This work is licensed under the Creative Commons BY-NC-ND 4.0 International License. Visit \url{https://creativecommons.org/licenses/by-nc-nd/4.0/} to view a copy of this license. For any use beyond those covered by this license, obtain permission by emailing \href{mailto:info@vldb.org}{info@vldb.org}. Copyright is held by the owner/author(s). Publication rights licensed to the VLDB Endowment. \\
\raggedright Proceedings of the VLDB Endowment, Vol. \vldbvolume, No. \vldbissue\ %
ISSN 2150-8097. \\
\href{https://doi.org/\vldbdoi}{doi:\vldbdoi} \\
}\addtocounter{footnote}{-1}\endgroup

\ifdefempty{\vldbavailabilityurl}{}{
\begingroup\small\noindent\raggedright\textbf{PVLDB Artifact Availability:}\\
The source code, data, and/or other artifacts have been made available at \url{\vldbavailabilityurl}.
\endgroup
}

%% file: sections_1_introduction_index.tex
\section{Introduction}

Data has become a central resource for scientific discovery and commercial innovation~\cite{hey2009fourth,Lazer2009CompSocialScience,sun2017revisiting}. As data commoditization accelerates, an increasing number of data marketplaces (e.g., Xignite, Snowflake, and Databricks) have emerged to facilitate the exchange and trading of datasets for model development across domains such as medicine, finance, and transportation~\cite{pei2020survey,liang2018survey,liu2021dealer,fernandez13data,chen2019towards}. Unlike traditional goods, however, the value of data is inherently context-dependent: a dataset is valuable only insofar as it improves downstream model performance. This creates a fundamental data management problem known as \emph{data valuation}: how to rigorously quantify, attribute, and compute the contribution of individual data items to model performance in an efficient and fair manner when multiple parties jointly participate in model training~\cite{liang2018survey,pei2020survey}.

The Shapley value~\cite{shapley1953value} provides a principled solution by assigning each data point its marginal contribution over all coalitions of the training data. It offers fairness guarantees and has become a standard foundation for data valuation~\cite{ghorbani2019data}\cite{lin2025comprehensive}. Yet exact Shapley computation is \#P-hard~\cite{deng1994complexity} because it requires evaluating an exponential number of subsets. Existing accelerations, including Monte Carlo sampling and truncated coalition evaluation~\cite{ghorbani2019data,Castro2009Polynomial}, influence- and gradient-based surrogates~\cite{basu2021influence,koh2017understanding}, and per-trajectory gradient attribution~\cite{wang2024data},
improve scalability but still rely on the \emph{global} coalition space or a fixed training trajectory, implicitly treating every training point as potentially relevant to every test point and overlooking the structural sparsity of individual predictions.

This assumption is overly pessimistic. Modern predictors exhibit strong structural sparsity: for a fixed test instance, only a limited portion of the training data participates in the computational pathway that determines the prediction~\cite{lin2022measuring,yang2023many,koh2017understanding,Ribeiro2016LIME,papyan2020prevalence}. For example, KNN predictions depend on nearby neighbors; decision trees depend on leaf-level partitions; kernel and margin-based models depend on active supports; graph neural networks depend on receptive fields~\cite{Lundberg2020TreeSHAP,Gilmer2017NeuralMessagePassing,cortes1995support,mairal2008supervised}. For a given test point, a large portion of training points are effectively irrelevant for that prediction. From a data management perspective, this reveals a structural opportunity: Shapley computation should exploit model-induced locality to avoid evaluating coalitions that cannot affect the outcome.

We formalize this structural sparsity as \textbf{model-induced locality}. For each test point, we define a \textit{support set} consisting of the training instances that influence its prediction through the model’s computational graph. When this \textbf{locality} is exact, projecting the coalition space onto the support region preserves the Shapley value. When locality is approximate, the deviation from global Shapley can be bounded by the aggregate influence of points outside the support. This reframes Shapley computation as a structured data processing problem: instead of enumerating $2^{|\mathcal D|}$ coalitions, we restrict attention to subsets of a much smaller support set.

However, locality alone is not sufficient. Even after restricting to supports, naïve computation remains exponential in the support size and suffers from extensive redundancy. The key insight of this paper is that the intrinsic complexity of Shapley valuation is governed not by the total number of coalitions, but by the number of \emph{distinct subsets that influence at least one valuation}. We prove that any correct algorithm must evaluate each such subset at least once, establishing an information-theoretic lower bound on retraining complexity. This perspective shifts the problem from exponential coalition enumeration to optimal subset reuse.

Guided by this principle, we propose \textbf{LSMR} (for \textbf{L}ocal \textbf{S}hapley via \textbf{M}odel \textbf{R}euse), a subset-centric exact algorithm. LSMR builds a bipartite support-mapping graph that links subsets to the training and test points whose utilities depend on them, and applies a pivot-based scheduling rule so that each distinct subset is trained exactly once. The resulting utility is propagated to all dependent valuations. LSMR removes both intra-support and inter-support redundancy and attains the intrinsic lower bound on retraining cost.

For larger support sets where exact enumeration remains expensive, we introduce \textbf{LSMR-A}, a reuse-aware Monte Carlo estimator. Instead of treating each sampled coalition independently, LSMR-A shares every sampled subset across all compatible support sets. The estimator remains unbiased and enjoys \emph{exponential concentration}, meaning that the probability of estimation error decreases exponentially fast with the number of samples. Importantly, its runtime depends on the number of \emph{distinct sampled subsets} rather than the total number of draws. This decouples sampling complexity from retraining complexity and reduces variance through amortized reuse. Under \emph{distribution shift}, when the test distribution differs from the training distribution and many training points become irrelevant to a given test instance, the variance advantage becomes \emph{structural}: irrelevant points are never sampled, so unnecessary randomness is eliminated at its source rather than merely reduced statistically.

We evaluate our framework across four representative model families, including weighted KNN, RBF Kernel SVM, decision trees, and graph neural networks, and diverse datasets. Experiments validate the theoretical claims: support-induced locality preserves valuation fidelity; LSMR-A achieves faster convergence and dramatically reduces retraining, matching the intrinsic subset complexity.

In summary, this paper makes four main contributions. We introduce model-induced locality as a structural abstraction for data valuation, formalizing support sets and deriving bounds on the gap between global and local Shapley values. We characterize the intrinsic subset complexity induced by these supports and establish an information-theoretic lower bound on the number of retraining operations required by any correct algorithm. Guided by this principle, we propose \textbf{LSMR}, an optimal subset-centric algorithm that trains each influential subset exactly once using support mapping and pivot scheduling. Finally, we develop \textbf{LSMR-A}, a reuse-aware Monte Carlo estimator that decouples sampling from retraining, remains unbiased with exponential concentration, and achieves lower variance through structural reuse.


%% file: sections_2_related_index.tex
\section{Related Work}\label{sec:related}
\paragraph{Data Valuation and Scalable Shapley Approximations.}
Data valuation~\cite{jiang2023opendataval, hammoudeh2024training} quantifies the contribution of individual training examples to model performance, supporting applications such as data attribution~\cite{chen2023algorithms}, federated learning~\cite{chen2023space}, model debugging~\cite{karlavs2023data}, and dataset compression~\cite{zhang2025prune}. Early leave-one-out methods~\cite{koh2017understanding} measure the effect of removing one example at a time but cannot capture interactions among data points. Shapley-value-based methods~\cite{ghorbani2019data,si2024counterfactual} address this limitation by aggregating marginal contributions over all coalitions, providing a cooperative-game-theoretic foundation for data valuation~\cite{shapley1953value}. Since exact Shapley computation is \#P-hard~\cite{deng1994complexity}, prior work has developed scalable approximations~\cite{pang2025shapley,luo2024fast}. Monte Carlo estimators~\cite{maleki2013bounding,wang2023data,ghorbani2019data,Castro2009Polynomial} sample random permutations, reinforcement-learning methods~\cite{yoon2020data} learn sampling policies, and structural shortcuts exploit convexity~\cite{jia2019towards}, submodularity~\cite{wei2015submodularity}, or influence-function reuse of gradients and Hessians~\cite{basu2021influence,pruthi2020estimating}. Recent methods also reduce or avoid retraining through amortization or trajectory-based approximation. Fast-DataShapley~\cite{sun2026fast} and Learnable Data Shapley~\cite{li2025toward} train reusable neural predictors that map samples directly to Shapley values, while In-Run Data Shapley~\cite{wang2024data} accumulates gradient-based attribution along a single SGD trajectory using Taylor approximations. Despite these advances, existing methods either operate over the full global coalition space or bypass it through amortized approximations. They do not explicitly exploit model-induced sparsity in the prediction pathway of individual test points, nor do they characterize the intrinsic subset complexity induced by structural locality or establish lower bounds on the number of distinct retraining operations required for Shapley computation.

\paragraph{Locality-Aware Shapley Computation.}
To address the inefficiency of global coalition enumeration, several recent works explore locality-aware Shapley evaluation. For unweighted KNN classifiers, Jia et al.~\cite{jia2019efficient} observe that prediction at a test point depends primarily on its nearest neighbors and show that ignoring distant points preserves relative importance rankings, enabling efficient approximate local Shapley computation. Wang et al.~\cite{wang2024efficient} extend this idea to hard-label KNN via dynamic programming, and Zhang et al.~\cite{zhang2025shapley} propose a near-linear-time weighted KNN-Shapley algorithm using a duplication method. While these methods highlight the practical benefits of restricting computation to local neighborhoods, their notion of locality is geometric and specific to KNN models. They do not generalize beyond distance-based neighborhoods, nor do they formalize locality as a structural property of the computational pathway of a model. In contrast, we introduce \emph{model-induced support sets}, which abstract locality directly from the architecture and prediction mechanism of the model. This formulation applies broadly to margin-based models, decision trees, kernel methods, and graph neural networks. We further characterize the intrinsic family of distinct subsets induced by these supports and prove that any correct algorithm must evaluate each such subset at least once. By exploiting model-induced locality and subset reuse, our methods enable efficient Shapley valuation while preserving strong theoretical guarantees.

%% file: sections_3_preliminary_index.tex
\section{Local Shapley}\label{sec:preliminary}
We first review the standard Shapley value for data valuation, then introduce a localized variant that restricts computation to training points relevant to a given test prediction. Finally, we discuss how this locality notion extends across different model families. A complete notation reference is provided in Supplemental Material~A.1.

\input{sections_3_preliminary_a_shapley}
\input{sections_3_preliminary_b_locality}
\input{sections_3_preliminary_c_model}

%% file: sections_3_preliminary_a_shapley.tex
\subsection{Shapley Value for Data Valuation}

Let $\mathcal{D}$ denote a training dataset, where each data point
$z = (\mathbf{x}_z, y_z)$ consists of a feature vector $\mathbf{x}_z$ and a class label $y_z$.
Data valuation~\cite{ghorbani2019data} seeks to quantify how much each
training point contributes to predictive performance on a testing set
$\mathcal{T}$. Formally, we aim to construct a valuation function
$\phi : \mathcal{D} \to \mathbb{R}$ that assigns an importance score to every
$z \in \mathcal{D}$.

\paragraph{Utility induced by retraining.}
For each test point $t \in \mathcal{T}$, we define a utility function $v_t : 2^{\mathcal{D}} \rightarrow \mathbb{R}$, where $v_t(S)$ measures the performance of a model trained on a subset $S \subseteq \mathcal{D}$ when evaluated at $t$. Throughout the paper, we adopt the standard retraining-based formulation as follows
\begingroup\small
\begin{equation}
v_t(S) \;=\; g_t\!\big(\theta(S)\big),
\end{equation}
\endgroup
where $\theta(S)$ denotes the model parameters obtained by training on $S$, and $g_t(\cdot)$ is a scalar evaluation functional at test point $t$. Examples of scalar evaluation functions include prediction confidence~\cite{jia2019efficient}, signed margin~\cite{koh2017understanding}, and loss reduction~\cite{ghorbani2019data}. We assume that model hyperparameters and training procedures are fixed across coalitions, so that variation in $v_t(S)$ arises solely from the choice of training subset $S$~\cite{ghorbani2019data,jia2019efficient,wang2023thresholdknnshapleylineartimeprivacyfriendly}. In the computational model considered throughout this work, the dominant cost of evaluating $v_t(S)$ is the training step required to obtain $\theta(S)$.

\paragraph{Global Shapley value.}
The Shapley value assigns to each training point its average marginal contribution over all possible coalitions~\cite{shapley1953value,chalkiadakis2011computational}.
For a fixed test point $t$ and training point $z \in \mathcal{D}$,
the Shapley value is
\begingroup\small
\begin{equation}
\phi_{z}(v_t)
=
\frac{1}{|\mathcal{D}|}
\sum_{S \subseteq \mathcal{D} \setminus \{z\}}
\frac{
v_t(S \cup \{z\}) - v_t(S)
}{
\binom{|\mathcal{D}|-1}{|S|}
}.
\end{equation}
\endgroup
Equivalently, $\phi_z(v_t)$ is the expected marginal gain of $z$ under a uniformly random permutation of $\mathcal{D}$, and it satisfies the classical axioms of efficiency, symmetry, null player, and additivity~\cite{shapley1953value}.

When multiple test points are considered, we aggregate per-test-point
values additively:
\begingroup\small
\begin{equation}
\phi_{z}
=
\sum_{t \in \mathcal{T}} \phi_{z}(v_t).
\end{equation}
\endgroup
This summation corresponds to evaluating total contribution over
$\mathcal{T}$.
By linearity of the Shapley value~\cite{luo_shapley_2022}, analysis can
be conducted independently for each test point and combined afterward.
Accordingly, we focus on a single fixed $t$ in the sequel.

\paragraph{Computational barrier and structural redundancy.}
Computing $\phi_z(v_t)$ requires evaluating marginal contributions over
all $2^{|\mathcal{D}|}$ subsets of $\mathcal{D}$.
Since each evaluation involves retraining the model on $S$ to obtain
$\theta(S)$, exact computation is computationally prohibitive.
In fact, computing the Shapley value of a single player in a general
cooperative game is \#P-hard~\cite{deng1994complexity}.

Crucially, the classical formulation treats every training point as potentially influential for every test point across all coalitions. That is, the global coalition space $2^{\mathcal{D}}$ implicitly assumes that influence may propagate arbitrarily through retraining. However, modern learning architectures often exhibit strong structural sparsity: for a fixed test point $t$, only a small portion of the training data meaningfully participates in the computational pathway that determines $v_t(S)$. This observation suggests that the global coalition space contains substantial structural redundancy. The next subsection formalizes this idea by introducing a localized utility that restricts attention to the subset of training points relevant to $t$, leading to the notion of \emph{Local Shapley value}.

%% file: sections_3_preliminary_b_locality.tex
\subsection{Local Shapley Value}
\label{sec:local-shapley}

Modern learning systems often exhibit strong structural
locality~\cite{lin2022measuring,yang2023many,koh2017understanding,Ribeiro2016LIME,papyan2020prevalence}: for a fixed test point $t$, only
a limited portion of the training data participates in the computational
pathway that determines $v_t(S)$.
This observation suggests that the global coalition space
$2^{\mathcal D}$ contains substantial redundancy when evaluating the
contribution of training data to $t$.

\paragraph{Support sets as structural locality.}
To formalize this idea, we introduce a \emph{support set} $\mathcal N(t) \subseteq \mathcal D$, defined as the subset of training points that can meaningfully influence the prediction or utility at $t$. The support set may be determined by architectural structure (e.g., receptive fields in GNNs, leaf membership in trees), margin sparsity (e.g., support vectors), kernel bandwidth, or other model-specific mechanisms. We treat $\mathcal N(t)$ as fixed with respect to a reference model trained once on the full dataset, so that locality reflects structural dependency rather than coalition-dependent drift. Its size can be controlled via model hyperparameters (e.g., $K$ in KNN~\cite{cover1967nearest}, threshold in kernel methods~\cite{parzen1962estimation}, hop radius in GNN~\cite{kipf2017gcn}), thereby trading fidelity for efficiency.

We define the \emph{projected (local) utility} as follows.
\begingroup\small
\begin{equation}
\label{eq:projected-utility}
v^{\mathcal{N}}_t(S)
:=
v_t(S \cap \mathcal N(t)),
\qquad
\forall S \subseteq \mathcal D.
\end{equation}
\endgroup
Since $v^{\mathcal N}_t(S)$ depends only on $S \cap \mathcal N(t)$, coalitions that agree on their intersection with $\mathcal N(t)$ are indistinguishable under the projected utility. Consequently, the cooperative game induced by $v^{\mathcal N}_t$ is effectively defined on the $2^{|\mathcal N(t)|}$ subsets of the support set, and players outside $\mathcal N(t)$ do not enlarge the intrinsic coalition space (formalized in Lemma~\ref{lem:local-equivalence}).

\begin{definition}[Local Shapley Value]
\label{def:local}
For a training point $z \in \mathcal D$, the local Shapley value with
respect to test point $t$ is defined as the classical Shapley value
computed under the projected utility $v^{\mathcal{N}}_t$:
\begingroup\small
\begin{equation}
\phi^{\mathrm{loc}}_z(v_t)
=
\frac{1}{|\mathcal D|}
\sum_{S \subseteq \mathcal D \setminus \{z\}}
\frac{
v^{\mathcal{N}}_t(S \cup \{z\}) - v^{\mathcal{N}}_t(S)
}{
\binom{|\mathcal D|-1}{|S|}
}.
\end{equation}
\endgroup
\end{definition}

From Eq.~\eqref{eq:projected-utility}, if $z \notin \mathcal N(t)$,
then $v^{\mathcal{N}}_t(S \cup \{z\}) = v^{\mathcal{N}}_t(S)$ for all $S$,
so $z$ acts as a null player and
$\phi^{\mathrm{loc}}_z(v_t)=0$.
For $z \in \mathcal N(t)$, the valuation depends only on coalitions
formed within the support set.

\paragraph{When does locality approximate the global game?}
Local projection modifies the cooperative game by removing
non-support players.
To quantify the resulting approximation error, we impose a stability
condition controlling how non-local points affect marginal contributions.

\begin{assumption}[Additive Non-local Stability]
\label{ass:nonlocal-stability}
Given a test point $t$, for any $z \in \mathcal N(t)$,
there exist nonnegative constants
$\{\ell_{t,z}(u)\}_{u \in \mathcal D \setminus \mathcal N(t)}$
such that for any
$S \subseteq \mathcal N(t)\setminus\{z\}$
and any
$U \subseteq \mathcal D \setminus \mathcal N(t)$,
\begingroup\small
\begin{equation*}
\bigl|
\Delta_z v_t(S \cup U)
-
\Delta_z v_t(S)
\bigr|
\le
\sum_{u \in U} \ell_{t,z}(u),
\end{equation*}
\endgroup
where
$\Delta_z v_t(S)
:=
v_t(S \cup \{z\}) - v_t(S)$.
\qed
\end{assumption}

This assumption formalizes a structural locality principle: non-support points influence the marginal effect of $z$ only weakly, and their aggregate impact scales at most additively~\cite{wang2024rethinking}.
This condition serves as the data-valuation counterpart to the classical algorithmic stability principle~\cite{bousquet2002stability,hardt2016train}. For stable learning algorithms—such as regularized ERM, kernel methods, or SGD on smooth objectives—including a single data point perturbs the trained parameters by an amount that decays monotonically with the coalition size $|S|$. This parameter stability directly guarantees that adding non-support instances cannot cause volatile swings in a player's marginal contribution, thereby ensuring the variations are strictly bounded.

\begin{proposition}[Approximation of Global Shapley]
\label{prop:approx-global}
\input{sections_a_proof_proof_prop_approx-global_claim}
\qed
\end{proposition}

Using the permutation interpretation of the Shapley value, a non support point influences the marginal contribution of \(z\) only when it appears before \(z\) in a uniformly random permutation, which happens with probability one half. Invoking Assumption~\ref{ass:nonlocal-stability} together with the bound established in the proposition above, the deviation between the global and local valuations is bounded by one half of the total non local interaction mass. The approximation error therefore depends only on the aggregate strength of interactions outside the support, rather than on the size of the full training set. Hence, whenever such effects decay with distance, kernel bandwidth, or graph hop radius, a moderately sized support set is sufficient to maintain valuation fidelity.

\textit{Example of regularized ERM.}
Consider $\ell_2$-regularized empirical risk minimization
$F_S(w)
=
\frac{1}{|S|}
\sum_{i \in S} f_i(w)
+
\frac{\lambda}{2}\|w\|_2^2$,
where each $f_i$ is convex and $L$-smooth. Then $F_S$ is $\lambda$-strongly convex, and classical uniform stability
analysis~\cite{bousquet2002stability} yields
$\|w_{S \cup \{u\}} - w_S\|_2
=
O\!\left(\frac{1}{\lambda |S|}\right)$.
If the evaluation functional $g_t(w)$ is $\psi$-Lipschitz in $w$, then
$|v_t(S \cup \{u\}) - v_t(S)|
=
O\!\left(\frac{\psi}{\lambda |S|}\right)$.
Moreover, the interaction term
$|\Delta_z v_t(S \cup \{u\}) - \Delta_z v_t(S)|
=
O\!\left(\frac{1}{\lambda |S|^2}\right)$
(up to alignment factors), which satisfies Assumption~\ref{ass:nonlocal-stability}, where $\ell_{t,z}(u)$ decays rapidly as shown in Proposition~\ref{prop:approx-global}. Consequently, the total non-local influence  $\sum_{u \notin \mathcal N(t)} \ell_{t,z}(u)$  remains small when interactions are weak.

\paragraph{Axiomatic properties.}
An important question is whether restricting the game to the support set preserves the fundamental fairness guarantees of the Shapley framework.

\begin{proposition}[Properties of Local Shapley]
\label{prop:axiomatic}
\input{sections_a_proof_proof_prop_axiomatic_claim}
\end{proposition}

Since $\phi^{\mathrm{loc}}_z(v_t)$ is the classical Shapley value applied to the projected utility $v^{\mathcal{N}}_t$, it satisfies symmetry, null-player, and additivity. Efficiency holds with respect to the projected utility $v^{\mathcal{N}}_t$; relative to the original utility $v_t$, any deviation from global efficiency corresponds precisely to the total utility mass discarded by excluding points outside $\mathcal N(t)$. When locality is exact (e.g., Threshold KNN~\cite{wang2023thresholdknnshapleylineartimeprivacyfriendly}), we have $v^{\mathcal{N}}_t = v_t$, and efficiency holds with respect to the original utility as well.

\paragraph{Discussion and limitations.}

The applicability of model-induced locality depends on the underlying architecture rather than being a guaranteed property of all learning systems. Highly non-convex or densely coupled architectures can exhibit global parameter dependencies where parameter shifts do not decay gracefully with coalition size, potentially widening the approximation gap in Proposition~\ref{prop:approx-global}.
Crucially, this limitation is manageable because the boundaries of the support set $\mathcal{N}(t)$ are inherently flexible. As detailed in Section~\ref{sec:model}, practitioners can systematically scale $\mathcal{N}(t)$ to capture long-range dependencies. Enlarging $\mathcal{N}(t)$ captures missing global interactions and tightens the theoretical approximation bound, offering a controllable trade-off between global fidelity and computational tractability.

%% file: sections_a_proof_proof_prop_approx-global_claim.tex
Under Assumption~\ref{ass:nonlocal-stability}, for any $z \in \mathcal N(t)$,
\begingroup\small
\begin{equation*}
\bigl|
\phi_z(v_t) - \phi^{\mathrm{loc}}_z(v_t)
\bigr|
\le
\frac{1}{2}
\sum_{u \in \mathcal D \setminus \mathcal N(t)}
\ell_{t,z}(u),
\end{equation*}
\endgroup
where $\phi_z(v_t)$ denotes the global Shapley value computed under the original utility $v_t$.

%% file: sections_a_proof_proof_prop_axiomatic_claim.tex
The local Shapley value satisfies: (i) \textbf{Symmetry}: identically contributing data points receive equal value; (ii) \textbf{Null Player}: a data point with zero marginal contribution in all coalitions receives zero value; and (iii) \textbf{Additivity}: the valuation is linear in $v_t$. 

%% file: sections_3_preliminary_c_model.tex
\subsection{Locality Induced by Model Architecture}
\label{sec:model}

Section~\ref{sec:local-shapley} introduced the notion of support set $\mathcal{N}(t)$ and showed that, whenever the prediction at $t$ depends only on this subset (exactly or approximately), the effective cooperative game collapses from $2^{|\mathcal D|}$ to $2^{|\mathcal N(t)|}$. We now demonstrate that such support sets arise naturally in common model families as a consequence of their computational structure. In some cases locality is exact and Local Shapley coincides with the global Shapley value; in others locality is approximate but satisfies Assumption~\ref{ass:nonlocal-stability}, yielding bounded error.

\begin{remark}[Sources of Locality]
In many widely used predictors, the prediction at $t$ primarily depends on a structurally determined subset $\mathcal{N}(t) \subseteq \mathcal D$:
\begin{enumerate}
    \item \textbf{KNN models}~\cite{cover1967nearest,dudani1976distance}: The prediction is computed solely from the $K$ nearest neighbors of $t$, so $\mathcal{N}(t)$ is naturally defined as this $K$-point neighborhood. Locality is determined by Euclidean distance and is exact under the threshold setting~\cite{wang2023thresholdknnshapleylineartimeprivacyfriendly}.
    \item \textbf{Support vector machines}~\cite{cortes1995support,chen2016distance}: The prediction admits a sparse dual form involving only support vectors, i.e., $\mathcal{N}(t)=\{z\in\mathcal D:\alpha_z\neq 0\}$. Locality is representational: although parameters are trained globally, only support vectors influence predictions.
    \item \textbf{Kernel methods with compact support}~\cite{parzen1962estimation,nadaraya1964estimating}: For kernels with finite bandwidth, only training points satisfying $\|x_z-x_t\|\le h$ contribute, yielding $\mathcal{N}(t)=\{z\in\mathcal D:\|x_z-x_t\|\le h\}$. Locality is governed by bandwidth.
    \item \textbf{Decision trees}~\cite{breiman1984classification,quinlan2014c4}: The prediction at $t$ is determined by the leaf reached along its decision path. The relevant training points are those passing through the same parent node, and we define $\mathcal{N}(t)=\{z\in\mathcal D:\; z \text{ reaches the same parent node as } t\}$. Locality arises from rule-based partitioning of the input space.
    \item \textbf{Graph neural networks}~\cite{kipf2017gcn,Gilmer2017NeuralMessagePassing,hamilton2017inductive}: In an $L$-layer message-passing GNN, only nodes within the $L$-hop receptive field of $t$ directly influence its embedding, so $\mathcal{N}(t)$ is the $L$-hop neighborhood of $t$. Locality is structural and induced by graph topology.
    \item \textbf{Deep neural networks}~\cite{bengio2013representation,papyan2020prevalence}: The learned encoder maps inputs into a representation space where semantically similar examples cluster together (e.g., via neural collapse). Consequently, $\mathcal{N}(t)$ is defined as the $K$-nearest training points to $t$ within this embedding space~\cite{wu2018unsupervised}.

\end{enumerate}
\label{remark:model}
\end{remark}

Across these families, locality emerges through distinct mechanisms: geometric proximity (KNN), dual sparsity (SVM), finite bandwidth (compact kernels), rule-based partitioning (trees), graph message passing (GNNs), and representational clustering (deep networks). In each case, $\mathcal{N}(t)$ is determined by the pathway---explicit in the architecture or implicit in the learned representation---that produces the prediction at $t$, and typically satisfies $|\mathcal{N}(t)|\ll|\mathcal D|$. When this dependency is strictly finite, Local Shapley recovers the global Shapley value exactly; when the influence decays but is nonzero, the approximation gap is governed by the aggregate non-local interaction mass characterized in Proposition~\ref{prop:approx-global}.

%% file: sections_4_exact_index.tex
\section{An Exact Algorithm for Local Shapley}
\label{sec:exact}

Building on the Local Shapley formulation in Section~\ref{sec:local-shapley}, we now turn to exact computation of local Shapley values. Although Definition~\ref{def:local} replaces the original utility with the projected utility $v_t^\mathcal{N}$, its combinatorial form still ranges over all $2^{|\mathcal D|}$ coalitions. We first prove that the projected utility induces an equivalent cooperative game whose effective player set is the support set $\mathcal N(t)$, thereby collapsing the intrinsic coalition space from $2^{|\mathcal D|}$ to $2^{|\mathcal N(t)|}$.

This reduction alone, however, does not eliminate computational redundancy. A naïve local evaluation still enumerates all subsets of $\mathcal N(t)$ separately for each training and test point. As shown in Figure~\ref{fig:reuse}, two redundancies and inefficiencies remain: \emph{intra-support redundancy}, where training points within the same support set require repeated evaluation of identical subsets, and \emph{inter-support redundancy}, where overlapping support sets across test points trigger duplicate subset retraining. In this section, we eliminate these redundancies in a structured manner: first by reorganizing the Shapley computation around subsets rather than players, and then by introducing a global reuse mechanism that guarantees each distinct subset is trained once while preserving exactness.

\input{sections_4_exact_figure_fig_reuse}

\input{sections_4_exact_a_baseline}

\input{sections_4_exact_b_reformulation}
\input{sections_4_exact_c_lsmr}
\input{sections_4_exact_d_complexity}

%% file: sections_4_exact_figure_fig_reuse.tex
\begin{figure}[t]
    \centering
    \subfigure[Intra-support redundancy]{
        \includegraphics[width=0.465\columnwidth]{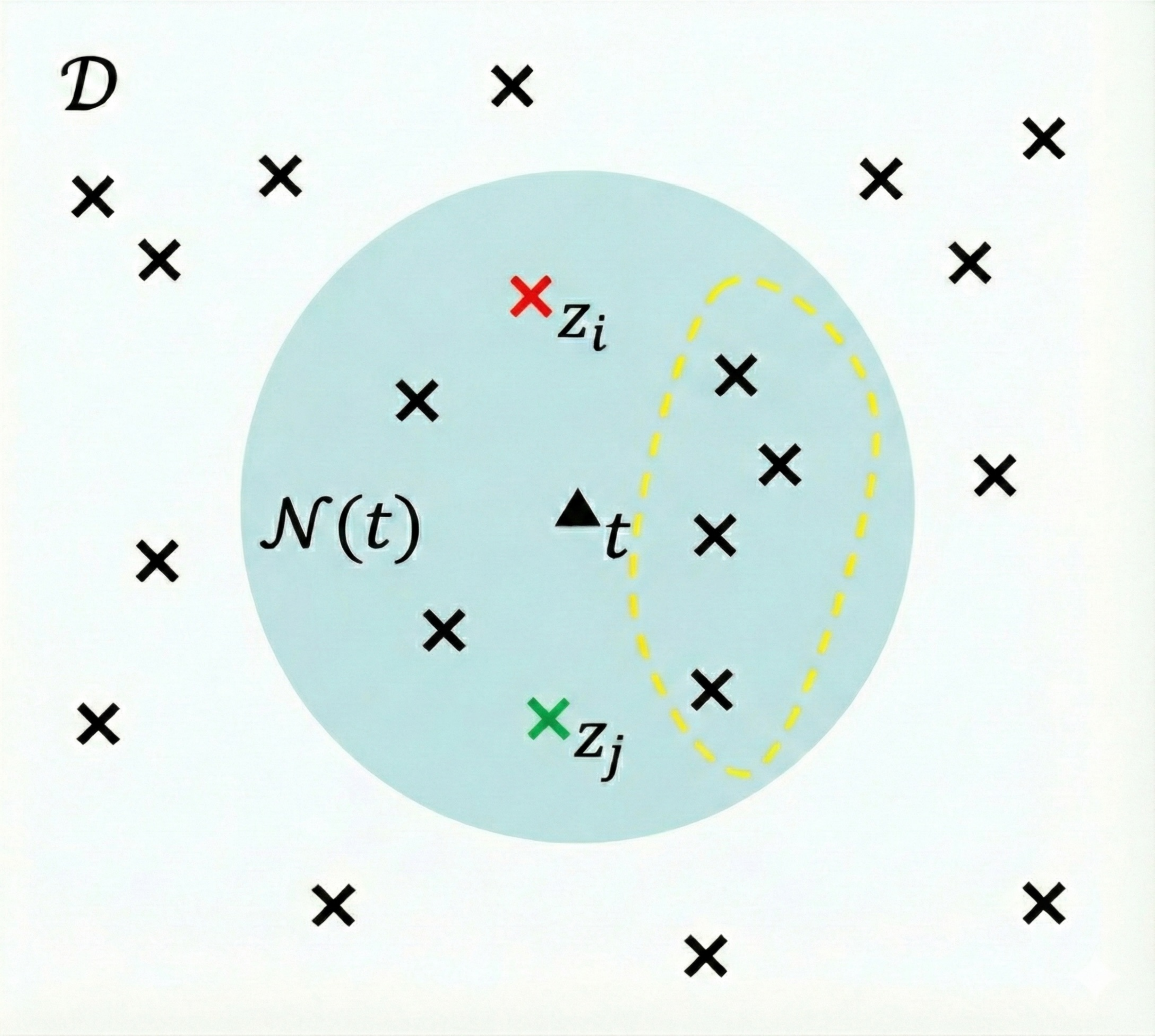}
    }
    \subfigure[Inter-support redundancy]{
        \includegraphics[width=0.465\columnwidth]{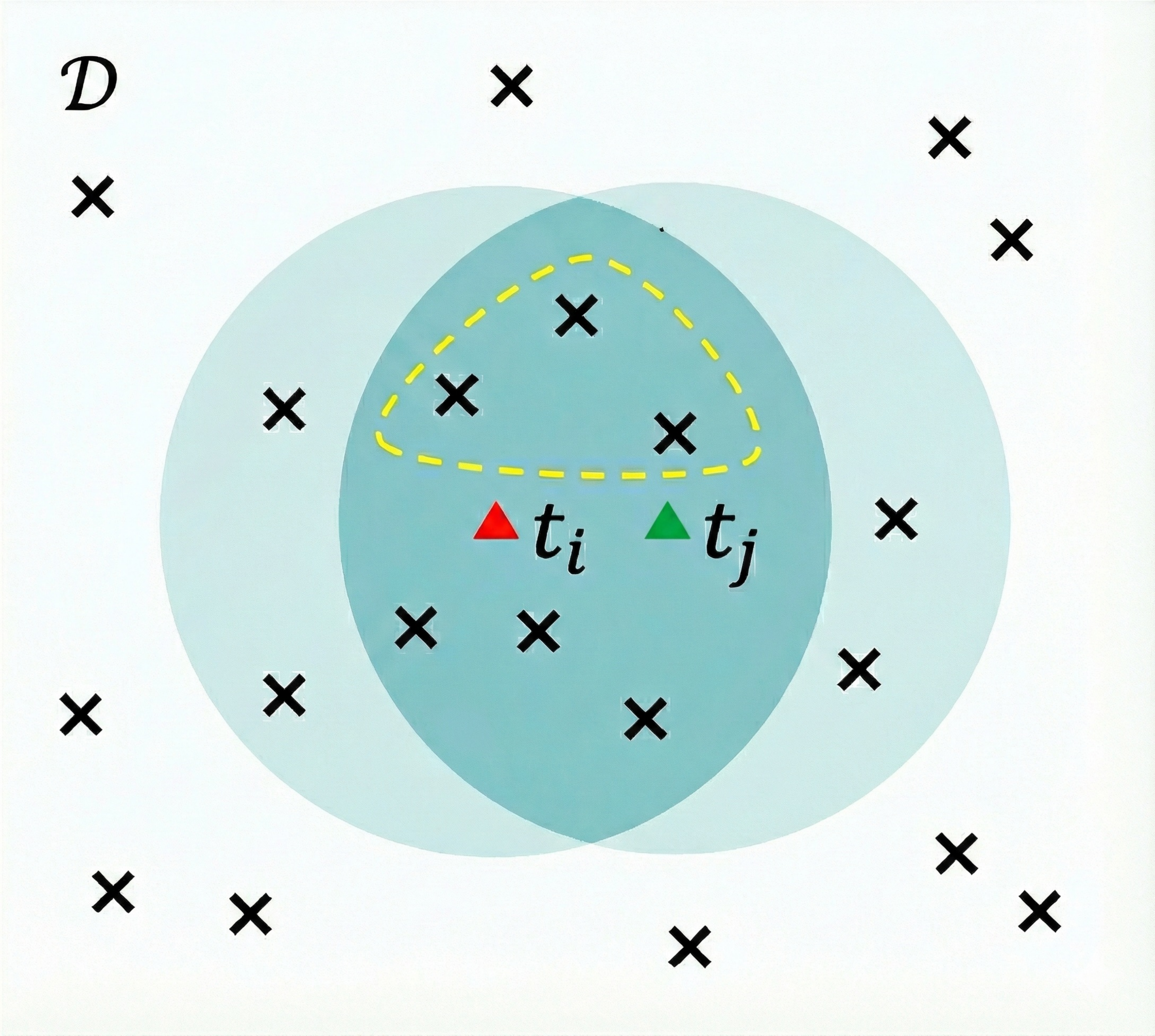}
    }
    \caption{(a) \textit{Intra-support redundancy}: the blue area denotes the support set $\mathcal{N}(t)$. For training points $z_i$ and $z_j$, many training subsets are shared in their Shapley value computations. (b) \textit{Inter-support redundancy}: test points $t_i$ and $t_j$ have overlapping supports (darker blue). Within this overlap, subsets are shared across their Shapley computations. The yellow dotted subset illustrates one such shared training subset.}
    \label{fig:reuse}
\end{figure}

%% file: sections_4_exact_a_baseline.tex
\subsection{Reducing the Game to the Support Set}
\label{sec:local-game}

We begin by formalizing the intrinsic dimensional reduction induced by the projected utility. Although Definition~\ref{def:local} is written over subsets of the full dataset $\mathcal D$, the projected utility satisfies $v_t^{\mathcal N}(S)=v_t(S\cap\mathcal N(t))$, implying that coalitions that agree on their intersection with $\mathcal N(t)$ are indistinguishable. Consequently, the Shapley computation for test point $t$ depends only on subsets of the support set $\mathcal N(t)$.

\begin{lemma}[Equivalence Under Projected Utility]
\label{lem:local-equivalence}
\input{sections_a_proof_proof_lem_local-equivalence_claim}
\end{lemma}
Lemma~\ref{lem:local-equivalence} shows that the Local Shapley value for $t$ can be computed by treating $\mathcal N(t)$ as the effective player set. Thus, the coalition space collapses from $2^{|\mathcal D|}$ to $2^{|\mathcal N(t)|}$. For subsets of size $k$, there are $\binom{|\mathcal N(t)|}{k}$ coalitions. As each marginal contribution requires evaluating $v(S)$ and $v(S \cup \{z\})$, enumeration over the reduced space yields the training cost as 
$2 \sum_{k=1}^{|\mathcal N(t)|} k \binom{|\mathcal N(t)|}{k}
= |\mathcal N(t)| \cdot 2^{|\mathcal N(t)|}$.
Aggregating over all test points, the total number of evaluations becomes
$\sum_{t \in \mathcal T} |\mathcal N(t)| \cdot 2^{|\mathcal N(t)|},$
which is much smaller than the $|\mathcal T|\cdot|\mathcal D|\cdot2^{|\mathcal D|}$ cost of global Shapley when $|\mathcal N(t)| \ll |\mathcal D|$. This serves as the natural local baseline and is summarized in Algorithm~\ref{alg:baseline}.

\input{sections_4_exact_algorithm_alg_baseline}

%% file: sections_a_proof_proof_lem_local-equivalence_claim.tex
The local Shapley value in Definition~\ref{def:local}, can be equivalently expressed as
\begingroup\small
\begin{equation*}
    \phi^{\mathrm{loc}}_z(v_t)
    =
    \frac{1}{|\mathcal{N}(t)|}
    \sum_{S \subseteq \mathcal{N}(t) \setminus \{z\}}
    \frac{
        v_t(S \cup \{z\}) - v_t(S)
    }{
        \binom{|\mathcal{N}(t)| - 1}{|S|}
    }.
\tag*{\qed}
\end{equation*}
\endgroup

%% file: sections_4_exact_algorithm_alg_baseline.tex
\begin{algorithm}[t]
\small
\caption{Baseline Algorithm}
\label{alg:baseline}
\begin{algorithmic}[1]
\Require Training set $\mathcal{D}$, test set $\mathcal{T}$
\Ensure Local Shapley value $\phi_z$ for each $z \in \mathcal{D}$
\For{each $z \in \mathcal{D}$}
    \State initialize $\phi_z \gets 0$
\EndFor
\For{each test point $t \in \mathcal{T}$}
    \State Retrieve support set $\mathcal{N}(t)$
    \For{each training point $z \in \mathcal{N}(t)$}
        \For{each subset $S \subseteq \mathcal{N}(t) \setminus \{z\}$}
            \State Train model $\theta(S)$ and evaluate $v_t(S)$;
            \State Train model $\theta(S \cup \{z\})$ and evaluate $v_t(S \cup \{z\})$;
            \State $\phi_z \gets \phi_z + 
            \frac{v_t(S \cup \{z\}) - v_t(S)}
            {|\mathcal{N}(t)| \cdot 
            \binom{|\mathcal{N}(t)| - 1}{|S|}}.$
        \EndFor
    \EndFor
\EndFor
\State \Return $\{\phi_z : z \in \mathcal{D}\}$
\end{algorithmic}
\end{algorithm}

%% file: sections_4_exact_b_reformulation.tex
\subsection{Subset-Centric Reformulation: Eliminating Intra-Support Redundancy}
\label{sec:subset-centric}

The baseline method (Algorithm 1) evaluates marginal contributions independently for each training point within the support set $\mathcal{N}(t)$. Thus, identical training subsets are repeatedly retrained when computing the marginal gains of different players within the same support set. To eliminate this intra-support redundancy, we break away from traditional player-centric loops and reorganize the valuation framework entirely around subsets. The key intuition is that a single utility evaluation $v_t(S)$ contains complete combinatorial information that can be distributed to all players in $\mathcal{N}(t)$ simultaneously, bypassing the need for separate marginal queries.

\begin{lemma}[Subset-Centric Reformulation]
\label{lem:local-closedform}
\input{sections_a_proof_proof_lem_local-closedform_claim}
\end{lemma}

Instead of computing differences ($v(S \cup \{z\}) - v(S)$), Lemma~\ref{lem:local-closedform} reformulates Local Shapley as a structured, weighted sum directly over the power set of the support region. The role of each subset $S$ is completely determined by whether the training point $z$ is contained within it. There are two cases: (i) \textbf{Positive Contributions ($z \in S$)}: Subsets containing $z$ correspond to valid coalitions \textbf{after} including the player, receiving a positive sign; and (ii) \textbf{Negative Contributions ($z \notin S$)}: Subsets lacking $z$ correspond to base coalitions \textbf{prior} to its addition, receiving a negative sign.

Crucially, evaluating $v_t(S)$ once allows us to update the values of all players in $\mathcal{N}(t)$ concurrently using these closed-form weights. This algorithmic shift yields an immediate computational dividend: instead of demanding $|\mathcal{N}(t)| \cdot 2^{|\mathcal{N}(t)|}$ utility evaluations per test point via player-centric loops, we evaluate only $2^{|\mathcal{N}(t)|}$ distinct local subsets. Intra-support redundancy is thus completely eradicated, dropping the computational overhead by an exact factor of $|\mathcal{N}(t)|$.


At this point, the computation is localized to $\mathcal N(t)$ and reorganized in a subset-centric manner, eliminating redundancy within each support set. Nevertheless, subsets that appear in overlapping supports of different test points are still evaluated independently. The next subsection addresses this remaining inefficiency by introducing a global reuse mechanism that ensures each distinct subset is trained only once across all test points.

%% file: sections_a_proof_proof_lem_local-closedform_claim.tex
For any test point $t$ and training point $z \in \mathcal{N}(t)$, the local Shapley value admits the following equivalent expression:
\begingroup\small
\begin{equation*}
\phi^{\mathrm{loc}}_{z}(v_t)
=
\frac{1}{|\mathcal{N}(t)|}
\sum_{S \subseteq \mathcal{N}(t)}
\frac{
(-1)^{\mathbb{I}(z \notin S)}\,
v_t(S)
}{
\binom{|\mathcal{N}(t)| - 1}{\,|S| - \mathbb{I}(z\in S)\,}
}.
\tag*{\qed}
\end{equation*}
\endgroup

%% file: sections_4_exact_c_lsmr.tex
\subsection{Global Subset Reuse Across Test Points}
\label{sec:lsmr}
\input{sections_4_exact_algorithm_alg_lsmr}

While subset-centric processing clears redundancies within an individual local game, nearby test points sampled from a shared data distribution still often exhibit heavily overlapping neighborhoods (supports). This overlap triggers duplicate evaluations of identical subsets across independent test iterations, inflating runtime because each unique evaluation requires an expensive model retraining step ($\theta(S)$).

To eliminate this remaining redundancy, we introduce \textbf{Local Shapley via Model Reuse (LSMR)}, which ensures that every distinct subset $S$ is trained exactly once across the entire computation and reused wherever it is valid. The key principle is global subset canonicalization: each subset has a unique designated evaluator, and all other occurrences reuse its result. LSMR achieves this through three components: a bipartite support-mapping graph, reverse support-set indexing, and pivot-based scheduling.

\subsubsection{Global Support Structure}

We represent training–test dependencies using a bipartite structure $G = (\mathcal D, \mathcal T, \mathcal N, \mathcal R)$, where $\mathcal N(t) \subseteq \mathcal D$ is the support of test point $t$, and
\begingroup\small
\begin{equation}
    \mathcal R(z) = \{\, t \in \mathcal T : z \in \mathcal N(t) \,\}
\end{equation}
\endgroup
is the reverse map listing the test points whose supports contain training point $z$. The mappings $\mathcal N$ and $\mathcal R$ are relational inverses, and together they characterize all potential reuse opportunities across test points.

\subsubsection{Reverse Support-Set Indexing}

For any subset $S \subseteq \mathcal D$, only test points whose supports contain $S$ can reuse its evaluation. These test points are exactly
\begingroup\small
\begin{equation}
    \mathcal R_S = \bigcap_{z \in S} \mathcal R(z),
    \qquad 
    \mathcal R_\emptyset = \mathcal T.
\end{equation}
\endgroup
$\mathcal R_S$ contains all test points for which $S \subseteq \mathcal N(t)$. Computing $\mathcal R_S$ requires intersecting adjacency lists, whose cost is negligible compared to model retraining as supports are typically small and sparse.

\subsubsection{Pivot-Based Scheduling}

To eliminate redundant trainings, LSMR assigns each subset $S$ a canonical evaluator. Intuitively, the first test point (under a fixed ordering) whose support contains $S$ ``owns'' the training of $S$; all later test points reuse the result. Fix a global ordering $\Pi_{\mathcal T}$ over test points and define the pivot
\begingroup\small
\begin{equation}
    t^*(S) = \arg\min_{t' \in \mathcal R_S} \Pi_{\mathcal T}(t').
\end{equation}
\endgroup
Subset $S$ is trained only when processing its pivot $t^*(S)$. For any other $t' \in \mathcal R_S$, the subset has already been evaluated at its pivot and the result is reused. This rule ensures that each distinct subset is trained exactly once, eliminating all inter-support redundancy.

\subsubsection{Complete LSMR Procedure}

Combining subset-centric reformulation with global reuse yields a unified pipeline. For each test point $t$, LSMR enumerates all subsets $S \subseteq \mathcal N(t)$. When encountering $S$, the algorithm computes $\mathcal R_S$ and checks whether $t$ is the pivot $t^*(S)$. If so, it trains $\theta(S)$, evaluates $v_{t'}(S)$ for all $t' \in \mathcal R_S$ and redistributes the utility values to all $z \in \mathcal N(t')$ using the closed-form weights from Lemma~\ref{lem:local-closedform}; otherwise, the subset has already been processed and no retraining occurs.

By construction, LSMR removes both intra-support redundancy (via subset-centric reformulation) and inter-support redundancy (via global pivot reuse), ensuring that each distinct subset $S$ is trained exactly once while preserving exact Local Shapley values.

%% file: sections_4_exact_algorithm_alg_lsmr.tex
\begin{algorithm}[t!]
\small
\caption{LSMR}
\label{alg:lsmr}
\begin{algorithmic}[1]
\State \textbf{Input:} Bipartite support-mapping structure $G = (\mathcal{D}, \mathcal{T}, \mathcal{N}, \mathcal{R})$, ordered test set $\Pi_{\mathcal{T}}$
\State \textbf{Output:} Local Shapley values $\phi_z$ for each $z \in \mathcal{D}$
\State Initialize $\phi_z \gets 0$ for all $z \in \mathcal{D}$
\ForAll{test point $t \in \mathcal{T}$ in order $\Pi_{\mathcal{T}}$}
    \State Retrieve support set $\mathcal{N}(t)$ from $G$
    \ForAll{subsets $S \subseteq \mathcal{N}(t)$}
        \If{$S = \emptyset$}\State $\mathcal{R}_S \gets \mathcal{T}$
        \Else
            \State $\mathcal{R}_S \gets \bigcap_{z \in S} \mathcal{R}(z)$
        \EndIf
        \State $t^* \gets$ first test point in $\mathcal{R}_S$ under $\Pi_{\mathcal{T}}$
        \If{$t = t^*$}
            \State Train model $\theta(S)$ and evaluate $v_{t'}(S)$ for all $t' \in \mathcal{R}_S$
            \ForAll{$t' \in \mathcal{R}_S$}
                \ForAll{$z \in \mathcal{N}(t')$}
                    \If{$z \in S$}
                        \State $\phi_{z} \mathrel{+}= \dfrac{v_{t'}(S)}{|\mathcal{N}(t')| \cdot \binom{|\mathcal{N}(t')| - 1}{|S| - 1}}$
                    \Else
                        \State $\phi_{z} \mathrel{-}= \dfrac{v_{t'}(S)}{|\mathcal{N}(t')| \cdot \binom{|\mathcal{N}(t')| - 1}{|S|}}$
                    \EndIf
                \EndFor
            \EndFor
        \EndIf
    \EndFor
\EndFor
\State \Return $\{\phi_z : z \in \mathcal{D}\}$
\end{algorithmic}
\end{algorithm}

%% file: sections_4_exact_d_complexity.tex
\subsection{Optimality and Complexity Analysis}
\label{sec:optimality}


We now quantify the computational savings of global subset reuse. Since the dominant cost is model retraining, we measure complexity by counting the number of distinct subsets $S$ for which $\theta(S)$ must be trained. Let
\begingroup\small
\begin{equation}
\mathcal S = \bigcup_{t \in \mathcal T} \{\, S \subseteq \mathcal N(t) \,\}
\end{equation}
\endgroup
denote the family of all distinct subsets that appear across support sets. The size $|\mathcal S|$ thus reflects the intrinsic retraining complexity of the Local Shapley problem under a given support mapping.

\begin{theorem}[Optimal Reuse]
\label{thm:maximal-reuse}
\input{sections_a_proof_proof_thm_maximal-reuse_claim}
\end{theorem}

Theorem~\ref{thm:maximal-reuse} shows that any correct algorithm must evaluate each subset in $\mathcal S$ at least once. Thus $|\mathcal S|$ is an information-theoretic lower bound on the number of model trainings. By construction, LSMR attains this bound exactly, performing one training per distinct subset and no more. By definition, we have
\begingroup\small
\begin{equation}
|\mathcal S|
\le
\min\!\left\{
\sum_{t \in \mathcal T} 2^{|\mathcal N(t)|},
\;
2^{|\mathcal D|}
\right\}.
\end{equation}
\endgroup
The first term corresponds to independently enumerating all subsets within each support set, while the second reflects the trivial upper bound over the entire dataset. In contrast, LSMR performs exactly $|\mathcal S|$ trainings, which can be substantially smaller when supports overlap. We therefore conclude across four levels of evaluation:
\begin{itemize}
\item \textbf{Global Shapley:} $|\mathcal T| \cdot |\mathcal D| \cdot 2^{|\mathcal D|}$ trainings.
\item \textbf{Local baseline:} $\sum_{t \in \mathcal T} |\mathcal N(t)| \cdot 2^{|\mathcal N(t)|}$ trainings.
\item \textbf{Subset-centric:} $\sum_{t \in \mathcal T} 2^{|\mathcal N(t)|}$ trainings.
\item \textbf{LSMR:} $|\mathcal S|$ trainings.
\end{itemize}
The final quantity depends only on the number of \emph{distinct} subsets induced by all supports, not on the number of test points or the number of players per support.

In practical regimes where test points are drawn from a fixed distribution, supports induced by nearby test points tend to overlap significantly. As $|\mathcal T|$ grows, many new test points introduce no new subsets, and the growth of $|\mathcal S|$ is sublinear relative to $\sum_{t \in \mathcal T} 2^{|\mathcal N(t)|}$. Consequently, the amortized number of model trainings per test point decreases with scale, making LSMR particularly effective in large evaluation settings.

\begin{corollary}[Amortized Reuse]
\label{cor:reuse-improves}
\input{sections_a_proof_proof_cor_reuse-improves_claim}
\qed
\end{corollary}

Therefore, the amortized number of model trainings per test point under LSMR vanishes as $T$ grows.




%% file: sections_a_proof_proof_thm_maximal-reuse_claim.tex
Let $\mathcal{R}_S$ be the set of test points whose support sets contain $S$. Any algorithm computing $\{v_t(S)\}_{t\in\mathcal{R}_S}$ must evaluate $v(S)$ at least once for every $S\in\mathcal S$.

%% file: sections_a_proof_proof_cor_reuse-improves_claim.tex
Assume the support mapping $\mathcal N(\cdot)$ has a finite range, i.e., only finitely many distinct support sets can appear across test points. Let $\mathcal S_T=\bigcup_{t\in\mathcal T_T}\{\,S\subseteq \mathcal N(t)\,\}$ denote the family of distinct subsets induced by the first $T$ test points. Then $|\mathcal S_T|$ is bounded as $T$ grows, and consequently $\frac{|\mathcal S_T|}{T}\to 0$ as $T\to\infty$. 

%% file: sections_5_approximation_index.tex
\section{Monte Carlo Approximation for Local Shapley}
\label{sec:approx}
While LSMR achieves the intrinsic retraining lower bound
and serves as the optimality reference for our analysis, its $2^{|\mathcal{N}(t)|}$
enumeration is impractical when supports are large. To tackle the scenarios with large supports, we
therefore propose \textbf{LSMR-A}, the reuse-aware Monte Carlo estimator developed
in this section. For stochastic approximation,  classical Monte Carlo methods~\cite{michalak2013efficient} estimate Shapley values by sampling permutations and averaging marginal contributions to avoid exhaustive enumeration. However, standard MC is \textit{player-centric}, retraining each sampled coalition independently and repeatedly, failing to exploit the structural reuse opportunities identified in Section~\ref{sec:lsmr}. This insight motivates a reuse-aware Monte Carlo strategy.

\subsection{From Exact Reuse to Reuse-Aware Sampling}
\label{sec:lsmra}

We therefore develop \textbf{LSMR-A}, a stochastic estimator that integrates the subset-centric formulation and pivot-based canonicalization of LSMR into the Monte Carlo setting. The resulting algorithm preserves unbiasedness and concentration guarantees while ensuring that each distinct subset is trained at most once across all samples and test points.

\subsubsection{Subset-Centric Monte Carlo Estimation}

Standard permutation-based MC estimates the Shapley value of a player $z$ by averaging marginal gains $v_t(S \cup \{z\}) - v_t(S)$ over sampled subsets $S$. This formulation ties each sample to a specific player, making cross-player reuse impossible. In contrast, Lemma~\ref{lem:local-closedform} expresses the Local Shapley value as a weighted expectation over subsets $S \subseteq \mathcal N(t)$. This observation enables a conceptual shift from a marginal-gain estimator to a \emph{subset-centric weighted utility estimator}. Instead of associating each sample with a single player, LSMR-A samples subsets $S$ and updates the Shapley values of \emph{all} players in $\mathcal N(t)$ simultaneously using the closed-form weights derived in Lemma~\ref{lem:local-closedform}. 

Under this formulation, a single model training on $S$ becomes a shared computational resource for the entire support set, eliminating intra-support redundancy in the stochastic setting.

\subsubsection{Pivoted Sampling for Cross-Test Reuse}

To extend reuse across test points, we integrate the pivot mechanism introduced in Section~\ref{sec:lsmr}. For each test point $t$, LSMR-A draws $M$ random permutations of $\mathcal N(t)$ and extracts the prefix-before-$t$ subset $S$ in the standard Shapley sampling manner.

For a sampled subset $S$, only test points whose supports contain $S$ can reuse its evaluation. These are precisely the elements of $\mathcal R_S$. To keep the estimator unbiased, we assign each subset a canonical evaluator via the pivot rule as follows.
\begingroup\small
\begin{equation}
t^*(S) = \arg\min_{t' \in \mathcal R_S} \Pi_{\mathcal T}(t').
\end{equation}
\endgroup

The sampling procedure follows two cases: (i) \textbf{Pivot hit:} If the current test point $t$ equals $t^*(S)$, the subset is accepted. The model $\theta(S)$ is trained once and reused to evaluate $v_{t'}(S)$ for all $t' \in \mathcal R_S$. These utilities are distributed to all $z \in \mathcal N(t')$ using the weighted subset-centric update rule; and (ii) \textbf{Pivot miss:} If $t \neq t^*(S)$, the sample is discarded. Since $S$ is assigned exclusively to its pivot $t^*(S)$, discarding avoids duplicate retraining without altering the sampling distribution.

This pivoted sampling scheme maximizes the reuse of trained models while preserving the unbiasedness of the estimator. The complete procedure is summarized in Algorithm~\ref{alg:lsmra}.

\input{sections_4_exact_algorithm_alg_lsmra}

\subsection{Statistical Guarantees}
\label{sec:lsmra-stats}
We first analyze the statistical validity and convergence properties of LSMR-A. The resulting guarantees mirror classical Monte Carlo analysis, while incorporating the effect of pivot-based reuse. We begin by establishing unbiasedness in expectation.

\begin{theorem}[Unbiasedness of LSMR-A]
\label{thm:lsmra-unbiased}
\input{sections_a_proof_proof_thm_lsmra-unbiased_claim}
\qed
\end{theorem}

The estimator remains unbiased because each subset $S$ is drawn from the same permutation-induced distribution as in classical MC, and the pivot rule merely canonicalizes which test point evaluates it. The weighted subset-centric update from Lemma~\ref{lem:local-closedform} preserves the exact expectation of the Local Shapley value.

We then prove exponential concentration of the estimator around the true Local Shapley value. The guarantee follows from independent permutation sampling together with standard bounded-difference inequalities.

\begin{theorem}[Concentration of LSMR-A]
\label{thm:lsmra-concentration}
\input{sections_a_proof_proof_thm_lsmra-concentration_claim}
\end{theorem}

The concentration bound immediately yields a sample complexity guarantee.

\begin{corollary}[Sample Complexity of LSMR-A]
\label{cor:lsmra-sample}
\input{sections_a_proof_proof_cor_lsmra-sample_claim}
\qed
\end{corollary}

Importantly, the number of samples $M$ required for $(\epsilon,\delta)$-accuracy matches classical MC. The distinction between LSMR-A and standard MC lies not in statistical efficiency, but in retraining efficiency.

\subsection{Computational Implications of Reuse}
\label{sec:lsmra-compute}

While $M$ determines statistical accuracy, runtime is governed by the number of \emph{distinct} subsets encountered during sampling, since each such subset is trained at most once.

\begin{theorem}[Expected Number of Distinct Sampled Subsets]
\label{thm:lsmra-distinct}
\input{sections_a_proof_proof_thm_lsmra-distinct_claim}
\qed
\end{theorem}

The sublinear growth in Theorem~\ref{thm:lsmra-distinct} reflects a saturation effect. Although many permutations are sampled, a large portion lead to overlapped subsets, so retraining increases far more slowly than the total sampling effort. Consequently, the effective number of model trainings is governed by the number of distinct subsets rather than by $M|\mathcal T|$. When supports overlap substantially, reuse dramatically reduces retraining cost while preserving statistical guarantees.

\subsection{Variance Reduction and Distribution Shift}
\label{sec:lsmra-variance}

Beyond correctness and runtime, reuse also improves estimator stability. Because utilities are shared across overlapping support regions, randomness that would otherwise be repeatedly injected in classical MC is amortized.

\begin{theorem}[Variance Reduction]
\label{thm:lsmra-variance}
\input{sections_a_proof_proof_thm_lsmra-variance_claim}
\end{theorem}
The variance reduction stems from removing conditional variance caused by redundant marginal evaluations. For the same sampling budget $M$, LSMR-A attains lower variance than classical MC.

This advantage becomes more pronounced under distribution shift. When the test distribution diverges from the training distribution, many training points lie outside the support set $\mathcal N(t)$ and have no influence on $v_t(\cdot)$. Classical MC nonetheless samples coalitions containing such irrelevant points, introducing unnecessary conditional variance. In contrast, LSMR-A restricts sampling to subsets of $\mathcal N(t)$ and therefore structurally removes this source of variance.

\begin{corollary}[Distribution Shift]
\label{cor:shift-gap}
\input{sections_a_proof_proof_cor_shift-gap_claim}
\qed
\end{corollary}

In summary, LSMR-A provides unbiased estimation with exponential convergence, retraining complexity determined by distinct subsets rather than the total number of samples, and strictly reduced variance under overlapping supports and distribution shift. 

%% file: sections_4_exact_algorithm_alg_lsmra.tex
\begin{algorithm}[t!]
\small
\caption{LSMR-A}
\label{alg:lsmra}
\begin{algorithmic}[1]
\State \textbf{Input:} Bipartite support-mapping structure $G = (\mathcal{D}, \mathcal{T}, \mathcal{N}, \mathcal{R})$, ordered test set $\Pi_{\mathcal{T}}$, number of Monte Carlo samples $M$
\State \textbf{Output:} Local Shapley values $\phi_z$ for each $z \in \mathcal{D}$
\State Initialize $\phi_z \gets 0$ for all $z \in \mathcal{D}$
\ForAll{test point $t \in \mathcal{T}$ in order $\Pi_{\mathcal{T}}$}
    \State Retrieve support set $\mathcal{N}(t)$ from $G$
    \For{$m = 1$ to $M$}
        \State Sample a random permutation $\pi$ of $\mathcal{N}(t) \cup \{t\}$
        \State $S \gets \{\pi_1, \dots, \pi_{i-1}\}$ where $i = \pi^{-1}(t)$
        \State Compute $\mathcal{R}_S$ and pivot $t^*$ as in Algorithm~\ref{alg:lsmr}, lines~7--11
        \If{$t = t^*$}
            \State Train model $\theta(S)$ and evaluate $v_{t'}(S)$ for all $t' \in \mathcal{R}_S$
            \For{all $t' \in \mathcal{R}_S$, $z \in \mathcal{N}(t')$} 
                \If{$z \in S$}
                    \State $\phi_z \mathrel{+}= \dfrac{(|\mathcal{N}(t')| + 1) \cdot v_{t'}(S)}{|S| \cdot M}$
                \Else
                    \State $\phi_z \mathrel{-}= \dfrac{(|\mathcal{N}(t')| + 1) \cdot v_{t'}(S)}{(|\mathcal{N}(t')| - |S|) \cdot M}$
                \EndIf
            \EndFor
        \EndIf
    \EndFor
\EndFor
\State \Return $\{\phi_z : z \in \mathcal{D}\}$
\end{algorithmic}
\end{algorithm}

%% file: sections_a_proof_proof_thm_lsmra-unbiased_claim.tex
In each Monte Carlo round, LSMR-A draws a uniform permutation of $\mathcal{N}(t)\cup\{t\}$ and takes the prefix-before-$t$ subset $S$. The pivot associated with $S$ is the first test point in the fixed ordering $\Pi_{\mathcal{T}}$, denoted $t^*(S)$. The model is trained on $S$ only when processing $t = t^*(S)$, and the resulting utilities $v_{t'}(S)$ are reused for all $t'$ such that $S \subseteq \mathcal{N}(t')$. Under this reuse mechanism, a MC estimator satisfies
\begingroup\small
\begin{equation*}
\mathbb{E}\!\left[ \widehat{\phi}(z) \right]
=
\sum_{t \in \mathcal{T}}
\frac{\mathbb{I}(z \in \mathcal{N}(t))}{|\mathcal{N}(t)|}
\sum_{S \subseteq \mathcal{N}(t)}
\frac{(-1)^{\mathbb{I}(z \notin S)}\, v_t(S)}
{\binom{|\mathcal{N}(t)| - 1}{\,|S| - \mathbb{I}(z \in S)\,}},
\end{equation*}
\endgroup
which coincides exactly with the closed-form Local Shapley value. Thus, LSMR-A is an unbiased estimator.

%% file: sections_a_proof_proof_thm_lsmra-concentration_claim.tex
Assume $|v_t(S)| \le B$ for all $t$ and all subsets $S$. Let $\widehat{\phi}(z)$ be the estimator obtained from $M$ samples. Then for any $\epsilon>0$,
\begingroup\small
\begin{equation*}
\Pr\!\left(
    \bigl|\widehat{\phi}(z) - \phi^{\mathrm{loc}}_z\bigr|
    > \epsilon
\right)
\le
2\exp\!\left(
    -\frac{2M\epsilon^2}{B^2}
\right).
\tag*{\qed}
\end{equation*}
\endgroup

%% file: sections_a_proof_proof_cor_lsmra-sample_claim.tex
To ensure
\begingroup\small
\begin{equation*}
\Pr\!\left(
    \bigl|\widehat{\phi}(z) - \phi^{\mathrm{loc}}_z\bigr| > \epsilon
\right)
\le \delta, 
\end{equation*}
\endgroup
it suffices to take
\begingroup\small
\begin{equation*}
M \;\ge\; \frac{B^2}{2\epsilon^2}\log\frac{2}{\delta}.
\end{equation*}
\endgroup
Moreover, due to subset reuse across test points, the effective number of required model trainings satisfies $M_{\mathrm{eff}} = \frac{M}{\mathbb{E}[\,|\mathcal{T}_S|\,]}$, where the expectation is taken over sampled subsets $S$.

%% file: sections_a_proof_proof_thm_lsmra-distinct_claim.tex
Fix a test point $t$ with support set $\mathcal{N}(t)$. Let $M$ rounds sample random permutations of $\mathcal{N}(t)\cup\{t\}$, and let $U_t$ be the collection of distinct prefix-before-$t$ subsets. Then, $\mathbb{E}\,|U_t| = O\!\left(\min\{2^{|\mathcal{N}(t)|}, \sqrt{M}\}\right)$.

%% file: sections_a_proof_proof_thm_lsmra-variance_claim.tex
Let $\widehat{\phi}^{\mathrm{MC}}(z)$ be the classical Monte Carlo estimator and $\widehat{\phi}^{\mathrm{LSMR\text{-}A}}(z)$ be the LSMR-A estimator. Then
\begingroup\small
\begin{equation*}
\begin{aligned}
\mathrm{Var}\!\left[\widehat{\phi}^{\mathrm{LSMR\text{-}A}}(z)\right]
&\;\le\;
\mathrm{Var}\!\left[\widehat{\phi}^{\mathrm{MC}}(z)\right]
-
\mathbb{E}\!\left[
    \mathrm{Var}\!\left(v_t(S)\mid S\right)
\right] \\
&\;\le\;
\mathrm{Var}\!\left[\widehat{\phi}^{\mathrm{MC}}(z)\right].
\end{aligned}
\tag*{\qed}
\end{equation*}
\endgroup

%% file: sections_a_proof_proof_cor_shift-gap_claim.tex
For a test point $t$ with support set $\mathcal{N}(t)$ and irrelevant points 
$\mathcal{D}_{\mathrm{irr}}=\mathcal{D}\setminus\mathcal{N}(t)$, the variance of the  Monte Carlo estimator decomposes as
\begingroup\small
\begin{equation*}
\begin{aligned}
\mathrm{Var}\!\left[\widehat{\phi}^{\mathrm{MC}}(z)\right]
&=
\mathrm{Var}\!\left[\widehat{\phi}^{\mathrm{MC}}(z)\mid S\subseteq\mathcal{N}(t)\right] \\
&\quad+
\mathrm{Var}\!\left(
    v_t(S\cup\{z\}) - v_t(S)
    \,\big|\,
    S\cap\mathcal{D}_{\mathrm{irr}}\neq\varnothing
\right).
\end{aligned}
\end{equation*}
\endgroup
Since LSMR-A samples only subsets of $\mathcal{N}(t)$, the second term is always zero for 
$\widehat{\phi}^{\mathrm{LSMR\text{-}A}}(z)$.

%% file: sections_6_experiment_index.tex
\section{Empirical Evaluation}
\label{sec:exp}

We conduct a comprehensive empirical evaluation of the Local Shapley framework across diverse model families. Our experiments are designed not only to assess predictive fidelity, but also to validate the structural and computational claims established in Sections~3–5, including intrinsic subset complexity, optimal reuse and retraining efficiency. We aim to answer five central research questions:

\begin{itemize}
    \item \textbf{RQ1 (Approximation Fidelity):} How accurately does Local Shapley approximate the global Shapley value across different model families, and under what conditions does structural locality preserve valuation quality?
    \item \textbf{RQ2 (Downstream Data Selection Utility):} Do valuations derived from Local Shapley remain effective for downstream tasks such as data selection, compared with global and alternative Shapley estimators?
    \item \textbf{RQ3 (Optimal Reuse for Efficiency):} How much does LSMR-A reduce model trainings and runtime compared to baselines, and how closely does its behavior match the intrinsic subset complexity in Theorem~\ref{thm:maximal-reuse}?
    \item \textbf{RQ4 (Sensitivity to Support Set Size):} How do Local Shapley fidelity and runtime change as the support size varies, and do the observed fidelity–runtime trade-offs agree with the theoretical analysis in Sections~\ref{sec:exact} and~\ref{sec:approx}?
    \item \textbf{RQ5 (Model-Induced Locality):} Does the support set need to align with the evaluated model? We construct support sets using one architecture (e.g., GNN), and evaluate utility under another (e.g., KNN), as discussed in Remark~\ref{remark:model}.
\end{itemize}

\subsection{Experimental Setup}
\label{sec:exp_setup}

\subsubsection{Locality Across Model Families and Datasets.}
To examine the generality of model-induced support sets, we instantiate locality across four representative model classes, each paired with a structurally aligned dataset: i) \textbf{Weighted $K$-Nearest Neighbors (WKNN)}~\cite{dudani1976distance} on \textbf{MNIST}~\cite{lecun1998gradient}, with $\mathcal{N}(t)$ defined as the $2K$ nearest neighbors in feature space; ii) \textbf{Decision Tree (DT)}~\cite{breiman1984classification} on \textbf{Iris}~\cite{fisher1936iris}, with $\mathcal{N}(t)$ consisting of training instances that share the same parent node as the leaf reached by $t$; iii) \textbf{RBF Kernel SVM (RBF-SVM) }~\cite{cortes1995support} on \textbf{Breast Cancer}~\cite{street1993nuclear}, with $\mathcal{N}(t) = \{ z \in \mathcal{D} : K(x_z, x_t) \ge 0.5 \}$; iv) \textbf{Graph Neural Network (GNN)}~\cite{kipf2017gcn} on \textbf{Cora}~\cite{mccallum2000automating}, with $\mathcal{N}(t)$ defined as the two-hop neighbors of $t$ under a two-layer GCN. These models span geometric proximity, kernel-induced decay, rule-based partitioning, and graph topology. For each model, the support set is constructed directly from its computational pathway, enabling measurement of support size and the associated retraining complexity. 

\subsubsection{Baselines.}
We compare LSMR-A against four representative Shapley estimation strategies spanning global evaluation, locality restriction, truncation, and complementary reformulation: i) \textbf{Global-MC}~\cite{jia2019towards}, which performs standard Monte Carlo estimation; ii) \textbf{Local-MC}, which restricts Monte Carlo sampling to the support set $\mathcal N(t)$ without reuse; iii) \textbf{TMC-S}~\cite{ghorbani2019data}, which applies truncated Monte Carlo with early stopping once predictions stabilize; iv) \textbf{Comple-S}~\cite{sun2024shapley}, which estimates Shapley values via complementary contribution reformulation. These baselines do not incorporate model-induced locality or reuse-based retraining reduction.

\subsubsection{Evaluation Metrics and Implementation Details.}
We evaluate two foundational aspects of Shapley value computation: i) \textbf{Fidelity} (correlation with global Shapley values and downstream task performance), ii) \textbf{Retraining Cost} (total running time and number of model trainings). Following~\cite{ghorbani2019data}, all Monte Carlo methods use the same stopping rule: every 100 samples, we test whether
\begingroup\small
\begin{equation}
\frac{1}{n}\sum_{i=1}^{n} \frac{|\phi_i^{(m)} - \phi_i^{(m-100)}|}{|\phi_i^{(m)}| + \epsilon} < \tau,
\label{eqn:converge}
\end{equation}
\endgroup
where $\phi_i^{(m)}$ denotes the estimate for training point $i$ after $m$ samples and $\epsilon = 10^{-12}$. $\tau=0.05$ controls convergence and serves as the stopping threshold for all models. Experiments were run on an Intel Xeon E5-2640 v4 CPU with 32 GB RAM. For GNN workloads, we used a single machine with 64 CPU cores and 8 NVIDIA H20 GPUs, running 64 jobs in parallel. Experimental details are provided in Supplemental Material~\ref{apx:exp}.

\subsection{RQ1: Approximation Fidelity}
\label{sec:exp_rq1}
\input{sections_6_experiment_figure_fig_rq1_correlation}
\input{sections_6_experiment_figure_fig_rq2_selection}

We first evaluate whether Local Shapley faithfully approximates the global Shapley value across model families. For each model–dataset pair, we compute global Shapley values using \textbf{Global-MC} and Local Shapley values using \textbf{LSMR-A} both run to convergence (Eq.~\eqref{eqn:converge}). Agreement is quantified using Pearson’s $r$ for linear consistency and Spearman’s $\rho$ for rank-order consistency. Across all four models, Local Shapley exhibits strong positive correlation with Global Shapley, with Pearson $r$ ranging from 0.516 to 0.839, indicating substantial agreement between local and global valuation. 

The strongest alignment occurs for \textbf{WKNN} (Figure~\ref{fig:corr_wknn_mnist}), where predictions are fully determined by the $K$ nearest neighbors and locality is highly concentrated. The strong correlation confirms that the support set captures the dominant influence pathway, consistent with the tight error bound in Proposition~\ref{prop:approx-global} when non-local interaction mass is negligible. For \textbf{Decision Tree} (Figure~\ref{fig:corr_dt_iris}) and \textbf{RBF-SVM} (Figure~\ref{fig:corr_svm_bc}), correlations remain consistently positive, reflecting that rule-based partitioning and kernel-induced decay capture the primary influence pathways. The slight reduction in correlation relative to WKNN aligns with approximate locality, as influence outside the support is nonzero but controlled.

For \textbf{GNN} (Figure~\ref{fig:corr_gnn_cora}), correlation is weaker but still clearly positive. This is expected: GNNs exhibit approximate locality, as nodes outside the $L$-hop receptive field may still affect learned parameters through shared gradients during training. Additionally, deep learning training introduces stochasticity via random initialization, mini-batch sampling, and non-convex optimization, which inflates variance in both the global reference and the local estimates. This empirical trend is consistent with the approximation bound in Proposition~\ref{prop:approx-global}: when non-local interaction mass is small but nonzero, correlation degrades gradually rather than collapsing.
We additionally compare against In-Run Data Shapley~\cite{wang2024data} on
GNN, the only applicable setting due to its reliance on gradient-based training (Supplemental Material~\ref{app:in-run-comparison}). While~\cite{wang2024data} completes valuation in a single training run, it yields lower approximation fidelity and data-selection accuracy. Conversely, LSMR-A preserves exact Shapley semantics while maintaining bounded error.

\subsection{RQ2: Downstream Data Selection Utility}
\label{sec:exp_rq2}

High correlation with global Shapley does not automatically imply downstream effectiveness. Since data valuation is commonly used for training set pruning and prioritization~\cite{ghorbani2019data,jia2019towards}, we evaluate whether Local Shapley preserves \emph{selection utility}. Specifically, we rank training points in descending order of their Shapley scores. Starting from an empty training set, we incrementally add top-ranked samples and retrain the model at each step. Test accuracy is plotted against the fraction of training data incorporated. A stronger valuation method should achieve high accuracy with fewer samples, producing a steeper and dominating selection curve. Across all four model pairs, LSMR-A consistently matches or exceeds the data selection performance of global estimators.

For \textbf{WKNN} (Figure~\ref{fig:sel_wknn_mnist}), where locality is strong, LSMR-A achieves the most improvement: approximately 10\% of locally selected data attains accuracy comparable to 20\% selected by Global-MC. This behavior is consistent with RQ1, where correlation is strongest under exact locality. For \textbf{Decision Tree} (Figure~\ref{fig:sel_dt_iris}), LSMR-A achieves the highest accuracy across most operating points, indicating that leaf-induced support sets preserve the ranking of influential samples even in small datasets.
For \textbf{RBF-SVM} (Figure~\ref{fig:sel_svm_bc}), LSMR-A and Local-MC perform comparably to Comple-S and outperform Global-MC and TMC-S in the low-data regime ($<10\%$). Although RQ1 shows only moderate linear correlation in this setting, the identification of the top-ranked samples remains robust—highlighting that data selection depends primarily on accurately identifying high-impact samples rather than perfectly matching global magnitudes. For \textbf{GNN} (Figure~\ref{fig:sel_gnn_cora}), LSMR-A yields the best selection performance. Global-MC performs worst, likely due to high retraining variance across coalitions in deep learning settings. In contrast, restricting valuation to 2-hop structural supports stabilizes ranking and better captures dominant influence pathways.

Overall, model-induced locality enables efficient analysis of data selection utility. The support-restricted approach identifies influential samples while substantially reducing retraining cost. As established in Theorem~\ref{thm:lsmra-unbiased} and ~\ref{thm:lsmra-concentration}, LSMR-A provides unbiased estimation with exponential concentration guarantees under permutation sampling. Moreover, Theorem~\ref{thm:lsmra-variance} shows that structured reuse further reduces estimator variance by amortizing redundant randomness across overlapping supports. Together, these results demonstrate that model-induced locality improves computational efficiency while preserving statistical stability.

\subsection{RQ3: Optimal Reuse for Efficiency}
\label{sec:exp_rq3}
\input{sections_6_experiment_table_tab_rq3}
In this section, we assess computational efficiency by measuring runtime and the number of model trainings required to satisfy the convergence criterion (Eq.~\eqref{eqn:converge}). Shown in Table~\ref{tab:rq3}, LSMR-A achieves the fastest convergence across all model families. The retraining counts show a consistent trend: on WKNN, LSMR-A reduces required trainings by more than three orders of magnitude compared to Global-MC, while on other models it delivers over $10\times$ speedups. The efficiency gains of LSMR-A arise from three mechanisms that eliminate redundant computation in Shapley evaluation: (i) coalition space reduction by restricting evaluation to $\mathcal{N}(t)$ (Lemma~\ref{lem:local-equivalence}); (ii) subset-centric reformulation to enable reuse within each support set (Lemma~\ref{lem:local-closedform}); and (iii) pivot scheduling to enable reuse across support sets (Theorem~\ref{thm:maximal-reuse}), achieving theoretically optimal reuse.

When supports are tight, as in \textbf{WKNN} where $|\mathcal{N}(t)| \ll |D|$, coalition space reduction alone already yields substantial gains: Local-MC achieves a significant speedup over Global-MC simply by restricting computation to the support set. LSMR-A further eliminates redundant evaluations during computation, reducing model trainings from $9.0\text{M}$ to $0.9\text{M}$. In contrast, for \textbf{RBF-SVM}, applying a kernel threshold of $0.5$ yields support sets that retain roughly $30\%$ of the training data, reducing the effectiveness of coalition space reduction through model-induced locality. Compared to Local-MC, which also restricts computation to the support set, LSMR-A substantially reduces training cost and converges more rapidly. The performance gap between the two highlights the importance of reuse: as support sets grow, limiting the coalition space alone becomes insufficient, and efficiency is primarily driven by sharing within and across support sets. Without carefully eliminating redundant subset evaluations through structured reuse, Local-MC even underperforms TMC-S and Comple-S in this setting. Detailed analyses of the efficiency contributions of each mechanism in LSMR-A are provided in the Supplemental Material~\ref{app:ablation}.
For \textbf{Decision Tree} and \textbf{GNN}, the speedup over Global-MC is relatively modest. Nevertheless, LSMR-A still achieves the lowest retraining count, indicating that the reuse mechanisms remain effective even under these less favorable structural conditions. 

\input{sections_6_experiment_figure_fig_scaling}

To further examine the scalability of LSMR-A, Figure~\ref{fig:scaling} reports the runtime and number of model trainings on the MNIST dataset as the training set size $|\mathcal{D}|$ increases, for two model families: WKNN ($|\mathcal{D}|=100$ to $10{,}000$, with test set size $|\mathcal{T}|=1{,}000$) and Decision Tree ($|\mathcal{D}|=50$ to $500$, $|\mathcal{T}|=100$). Note that the results of certain baselines are not reported once their runtime becomes prohibitively expensive, i.e., when they require more than 24 hours to converge. 

On \textbf{WKNN} (Figure~\ref{fig:scaling}(a)), all global baselines exhibit steep growth in both runtime and training count over their observable ranges. Local-MC flattens earlier due to coalition space reduction but still requires substantial model trainings. In contrast, LSMR-A remains nearly flat across the entire range. At $|\mathcal{D}|=10{,}000$, LSMR-A is more than five orders of magnitude faster than Global-MC, converging in under two minutes where global methods are infeasible to run.
On \textbf{Decision Tree} (Figure~\ref{fig:scaling}(b)), LSMR-A stays nearly constant on both metrics across the entire range, whereas Global-MC increases by more than two orders of magnitude as $|\mathcal{D}|$ grows from 50 to 200. Local-MC also levels off, but at a scale about one order of magnitude higher than LSMR-A, further demonstrating that the subset reuse mechanism consistently yields additional computational savings.

This scaling trend is consistent with Theorems~\ref{thm:lsmra-concentration} and~\ref{thm:lsmra-distinct}: when the support size is fixed, the number of distinct subsets is bounded independently of $|\mathcal{D}|$, so the amortized retraining cost per test point decreases as the dataset grows, with concentration guarantees preserving estimation stability under reuse. The widening gap between LSMR-A and global baselines across both model families further demonstrates that the framework remains effective in regimes where global methods are computationally infeasible.

\subsection{RQ4: Sensitivity to Support Set Size}
\label{sec:exp_rq4}
\input{sections_6_experiment_table_tab_rq4}

Table~\ref{tab:rq4} examines how the support size $|\mathcal N(t)|$ mediates the trade-off between approximation fidelity and retraining time cost. We use WKNN on MNIST and vary the size of $|\mathcal{N}(t)|$. All methods are evaluated at the common convergence point (Eq.~\eqref{eqn:converge}). To assess downstream utility, we introduce Acc@$x\%$, denoting the test accuracy achieved when training on the top-$x\%$ of selected data.

As $|\mathcal{N}(t)|$ increases, approximation fidelity improves, with Pearson correlation rising by approximately 103\% from the smallest to the largest support. Most gains are realized before $|\mathcal{N}(t)| = 10$, after which improvements plateau, consistent with Proposition~\ref{prop:approx-global}: the primary contributors enter the support early, and further expansion mainly reduces residual non-local effects with diminishing returns. Across all settings, LSMR-A remains more than three orders of magnitude faster than Global-MC. Although the coalition space scales as $2^{|\mathcal{N}(t)|}$, the number of model trainings and wall-clock time grow sublinearly, as both intra-support and inter-support reuse amortize computation across training and test points respectively.

Even with $|\mathcal N(t)|=3$, LSMR-A outperforms Global-MC in data selection by concentrating computation on outcome-relevant coalitions and producing more stable rankings.
As $|\mathcal N(t)|$ grows further, correlation with the global estimator continues to improve, but selection accuracy remains stable. This is expected: including more training points in the support set reduces the non-local interaction mass outside $\mathcal{N}(t)$, tightening the approximation bound in Proposition~\ref{prop:approx-global} and bringing the local game closer to the global game in magnitude. Data selection accuracy does not benefit that much from this continued improvement. This suggests that even a relatively small support set already captures the dominant influence pathway that determines the prediction at each test point, and thus suffices to identify the most valuable training instances. Enlarging the support beyond this point refines numerical agreement with the global estimator but does not change which training points are ranked highest, as the additional nodes contribute only marginally to the prediction and receive correspondingly small Shapley values.

\subsection{RQ5: Model-Induced Locality}
Table~\ref{tab:rq5} examines whether locality must align with the utility evaluation model. While earlier experiments construct $\mathcal{N}(t)$ and compute utility under the same model architecture, we introduce cross-model settings to isolate the effect of architectural alignment. We build $\mathcal{N}(t)$ using one model’s computational pathway (e.g., $K$-nearest neighbors for WKNN or same-leaf membership for Decision Tree) but evaluate utility under a different model, keeping the convergence criterion and average support size fixed. We also include a \textbf{Random} baseline that samples $|\mathcal{N}(t)|$ training points uniformly.

For WKNN, model-aligned supports deliver the highest correlation and strongest selection performance. Substituting Decision Tree pathways leads to a moderate decrease, indicating partial consistency between the two locality definitions, since feature-based splits can still cluster nearby samples. The reduction is greater with RBF-SVM supports, where a stringent kernel threshold produces sparse neighborhoods that misses WKNN’s locality structure. Even so, both cross-model supports markedly surpass the random baseline, with Pearson correlations more than five times higher. This suggests LSMR-A remains robust to architectural mismatch as long as the support set preserves a partial model prediction pathway.

For GNN, misalignment has a substantially stronger impact. Replacing graph-structural supports with WKNN or Decision Tree neighborhoods nearly eliminates correlation ($r$ and $\rho \approx 0.1$), even worse than random performance. This stems from a severe structural mismatch: GNN predictions depend on message passing over graph topology, while WKNN and Decision Tree supports are defined in feature space. Nodes that are close in feature space may be far apart in the graph and thus have minimal influence on GNN predictions, and vice versa. Consequently, the misaligned support set violats the assumption that $\mathcal{N}(t)$ contains the most relevant training points for $v_t$. In this case, the approximation bound in Proposition~\ref{prop:approx-global} no longer provides meaningful guarantees, as interactions outside the intended local region dominate the valuation.

These results show that locality is model-dependent. Aligning support set construction with the evaluation model’s computational pathway is essential for preserving valuation fidelity, and the penalty for misalignment grows with structural differences in their definitions of locality as shown in Remark~\ref{remark:model}.

\input{sections_6_experiment_table_tab_rq5}

%% file: sections_6_experiment_figure_fig_rq1_correlation.tex
\begin{figure*}[ht]
    \centering
    \subfigure[WKNN, MNIST]{
        \includegraphics[width=0.235\textwidth]{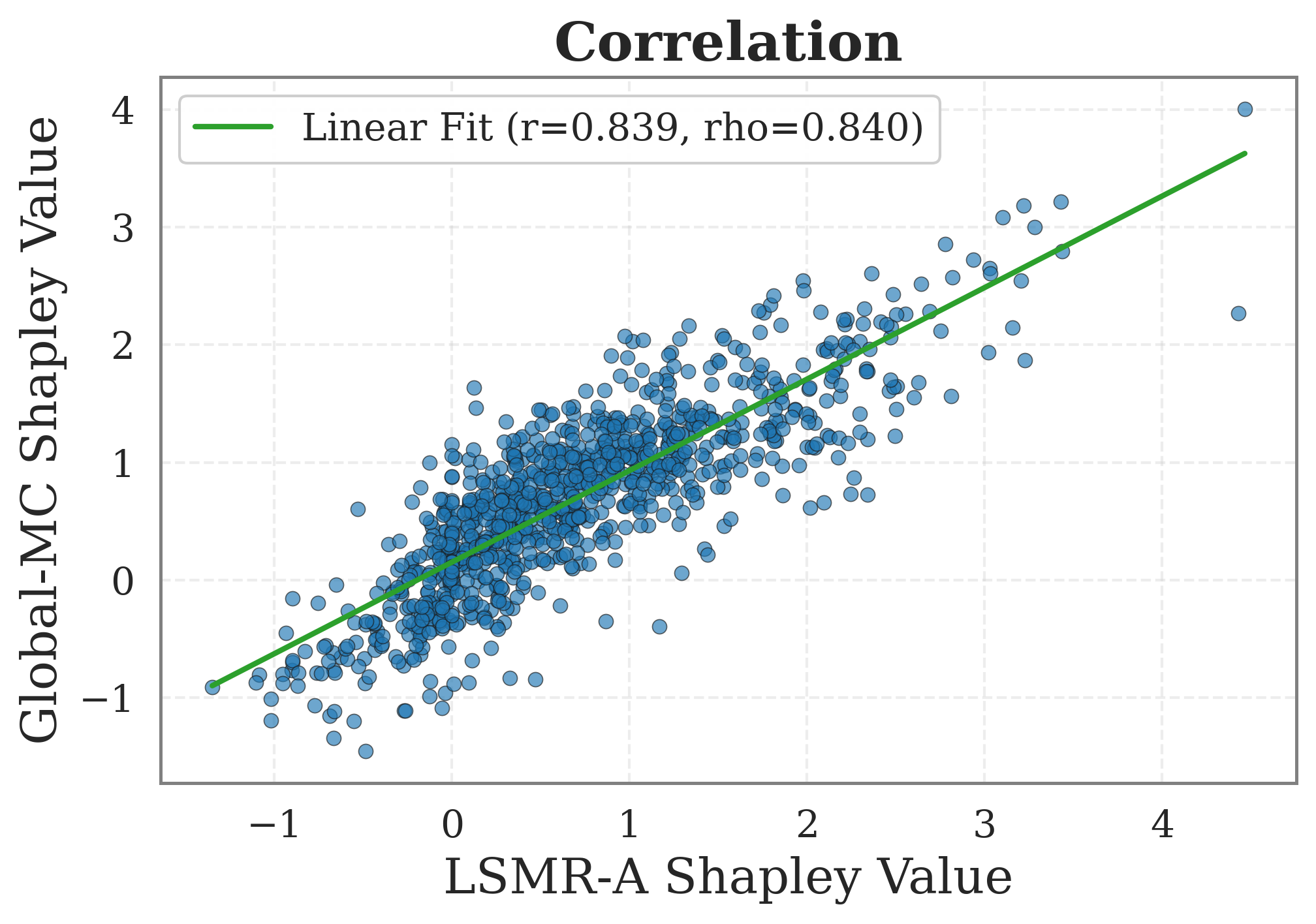}
        \label{fig:corr_wknn_mnist}
    }
    \subfigure[DT, Iris]{
        \includegraphics[width=0.235\textwidth]{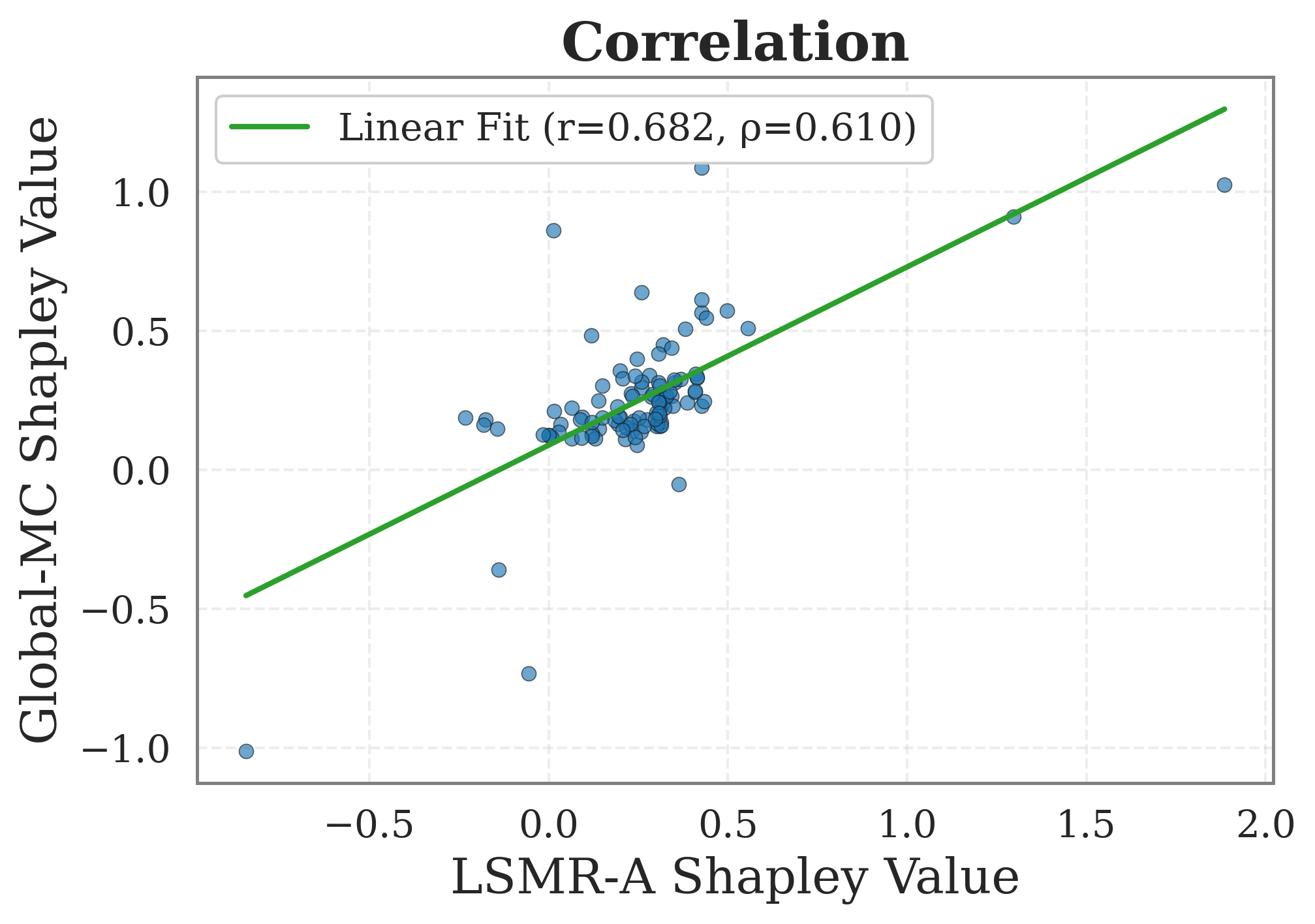}
        \label{fig:corr_dt_iris}
    }
    \subfigure[RBF-SVM, Breast Cancer]{
        \includegraphics[width=0.235\textwidth]{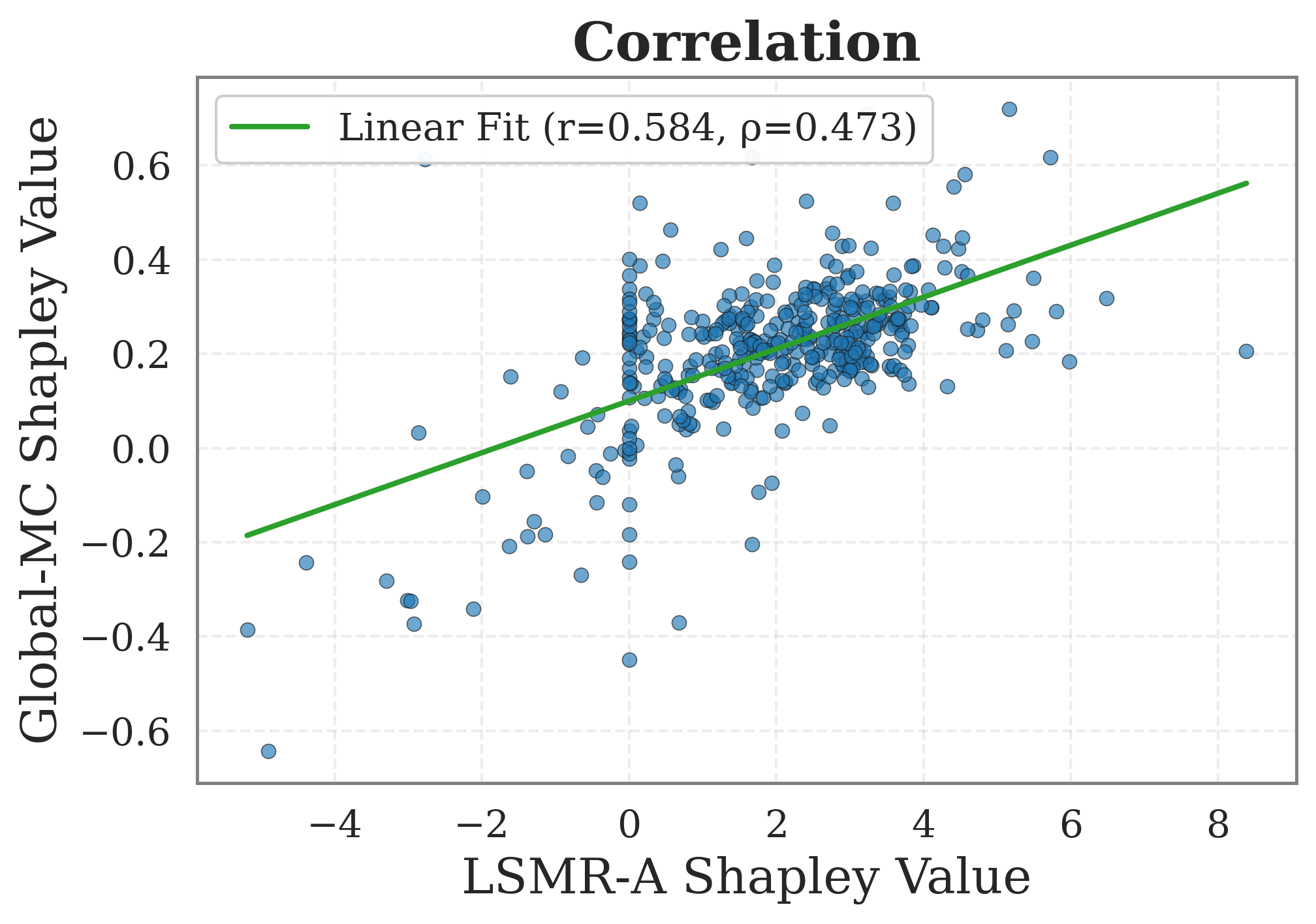}
        \label{fig:corr_svm_bc}
    }
    \subfigure[GNN, Cora]{
        \includegraphics[width=0.235\textwidth]{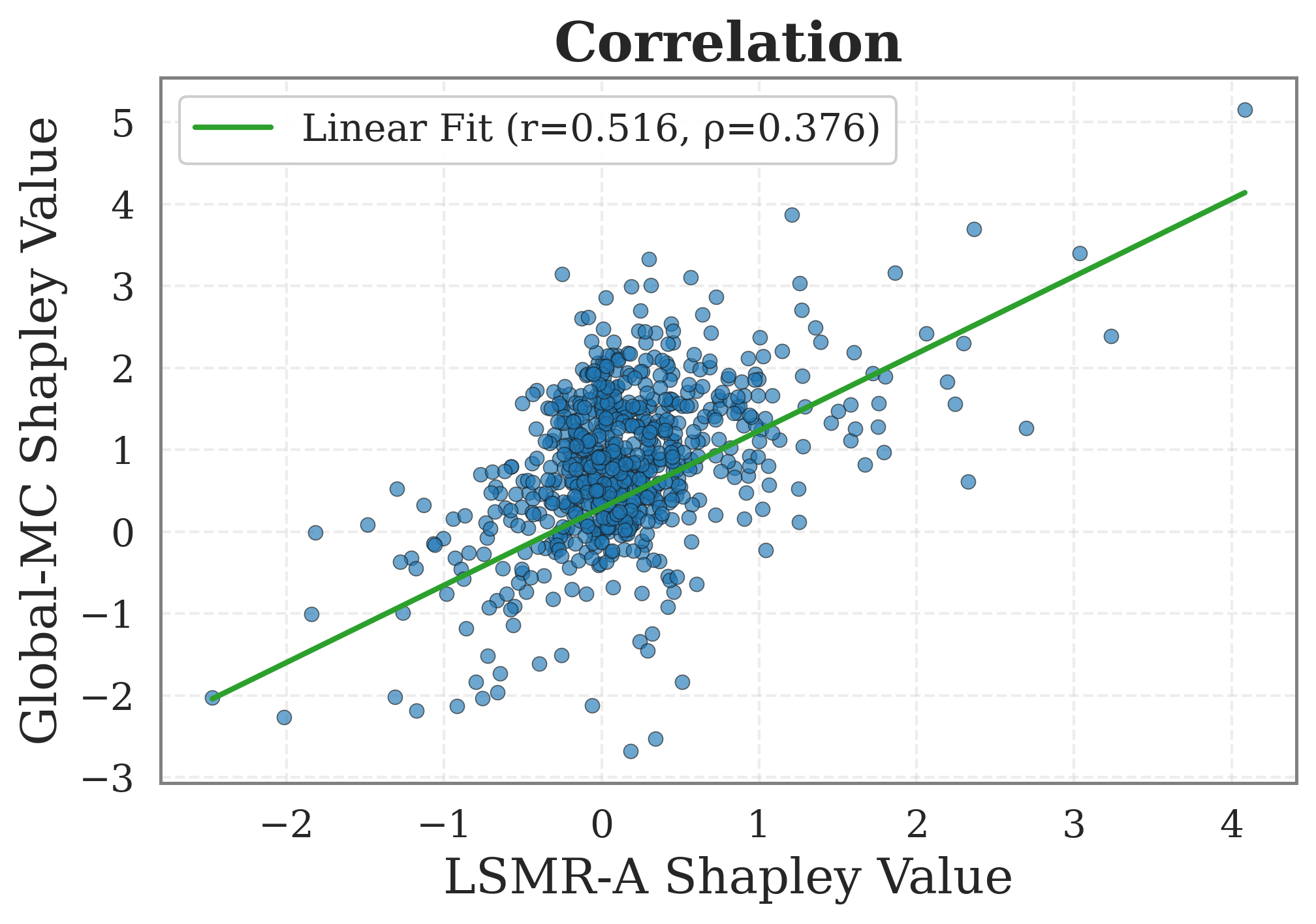}
        \label{fig:corr_gnn_cora}
    }
    \caption{Scatter plots of Local Shapley (x-axis) versus Global Shapley (y-axis). Green lines indicate linear regression fits. }
    \label{fig:rq1_correlation}
\end{figure*}

%% file: sections_6_experiment_figure_fig_rq2_selection.tex
\begin{figure*}[t]
    \centering
    \subfigure[WKNN, MNIST]{
        \includegraphics[width=0.235\textwidth]{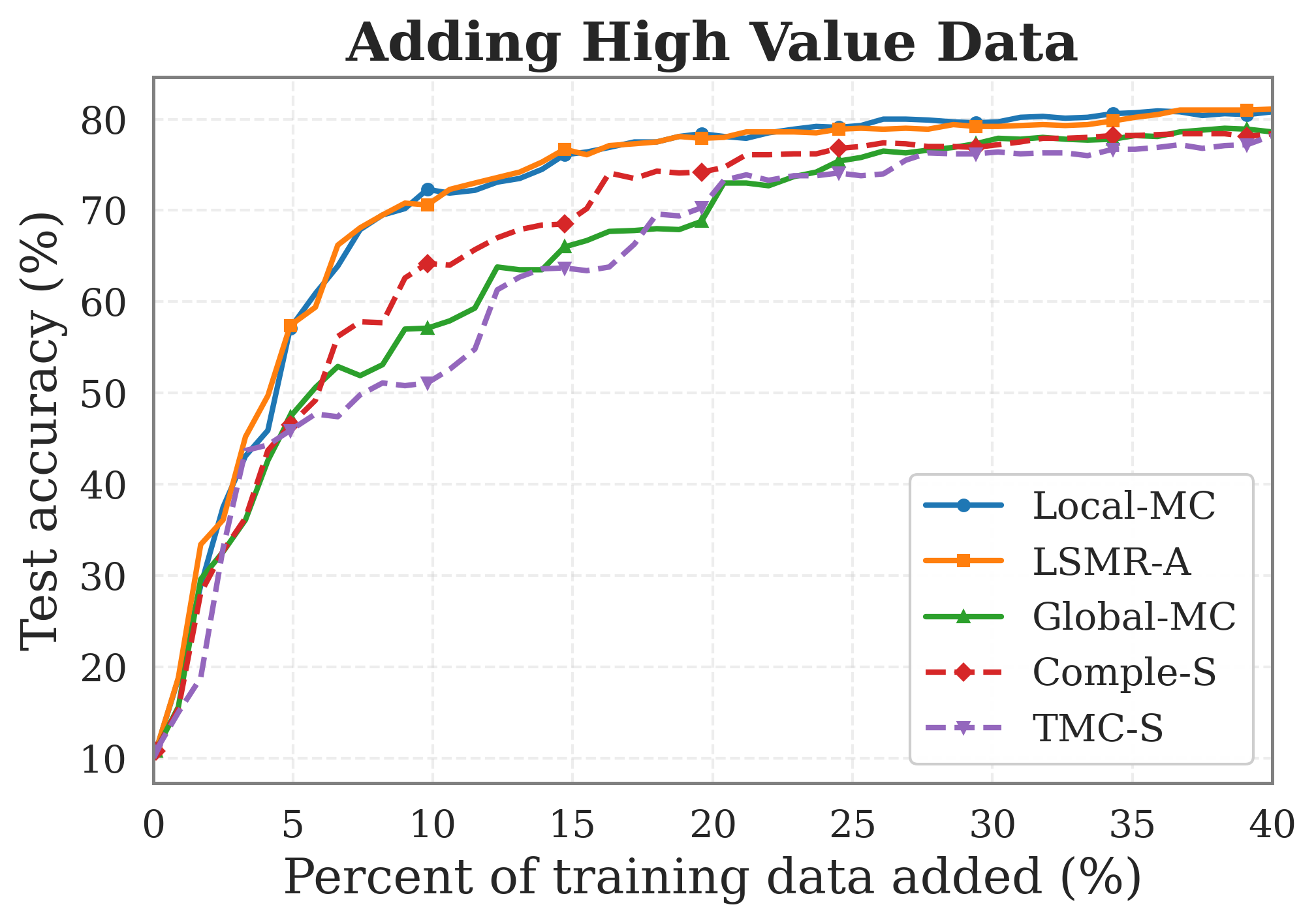}
        \label{fig:sel_wknn_mnist}
    }
    \subfigure[DT, Iris]{
        \includegraphics[width=0.235\textwidth]{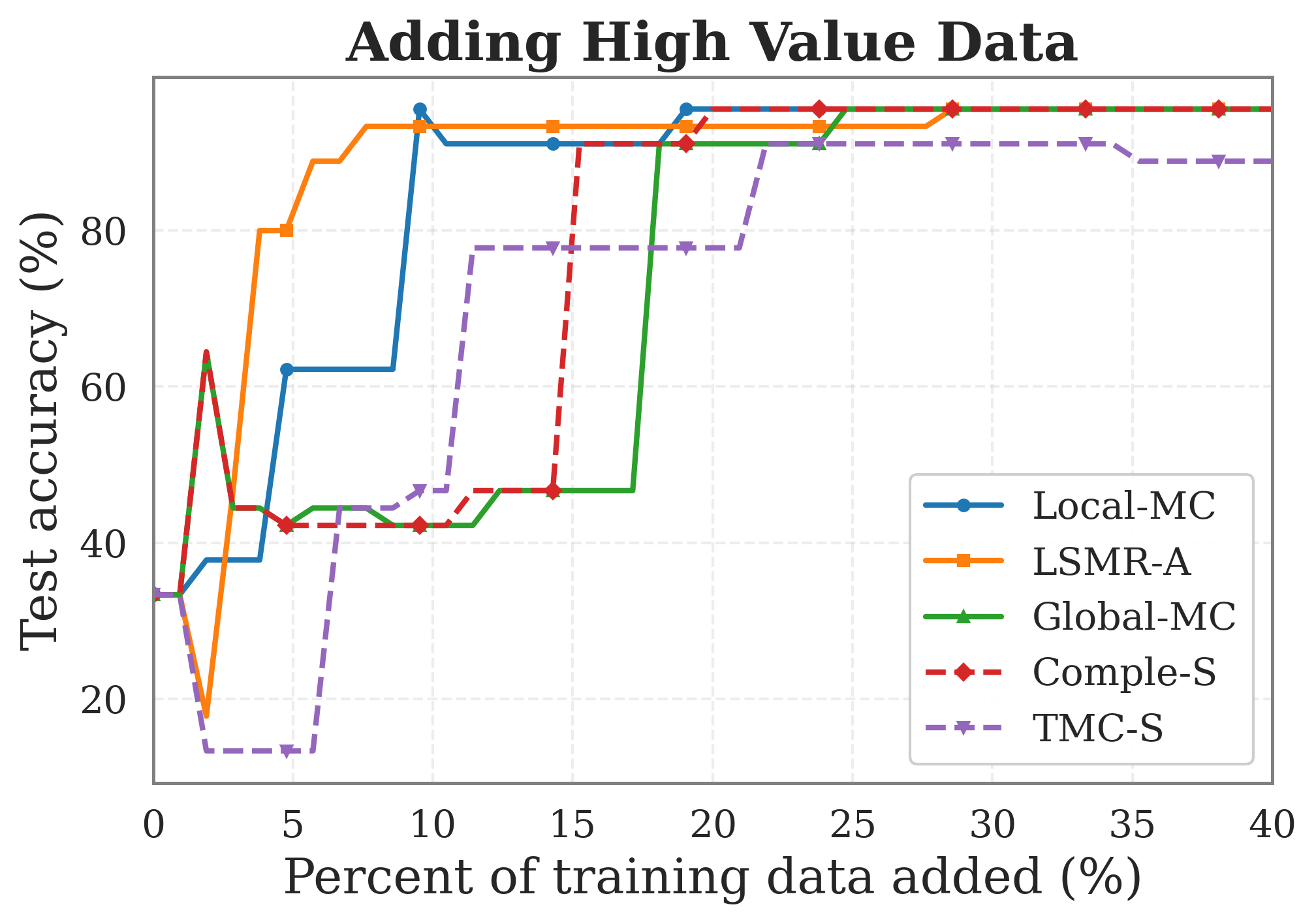}
        \label{fig:sel_dt_iris}
    }
    \subfigure[RBF-SVM, Breast Cancer]{
        \includegraphics[width=0.235\textwidth]{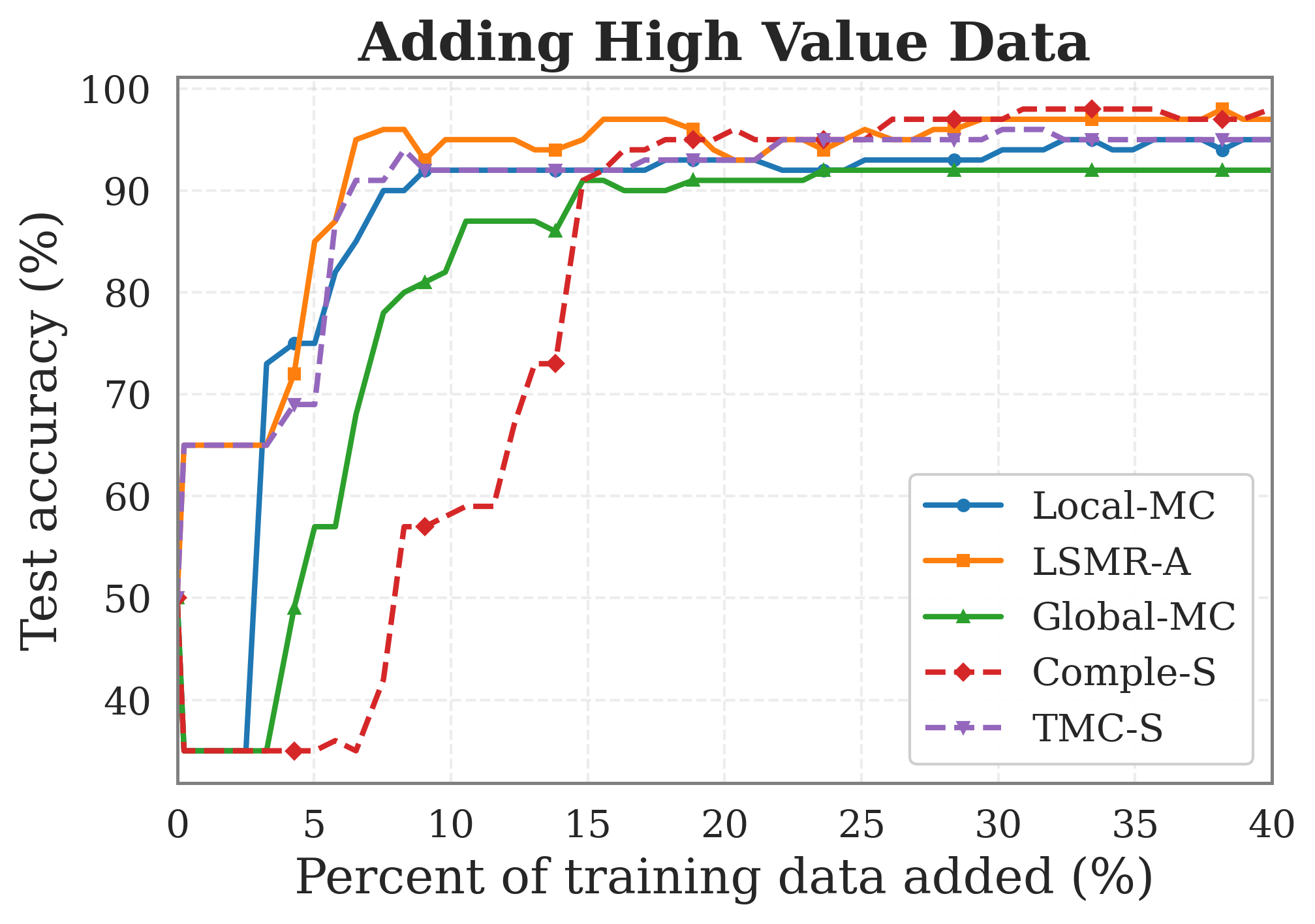}
        \label{fig:sel_svm_bc}
    }
    \subfigure[GNN, Cora]{
        \includegraphics[width=0.235\textwidth]{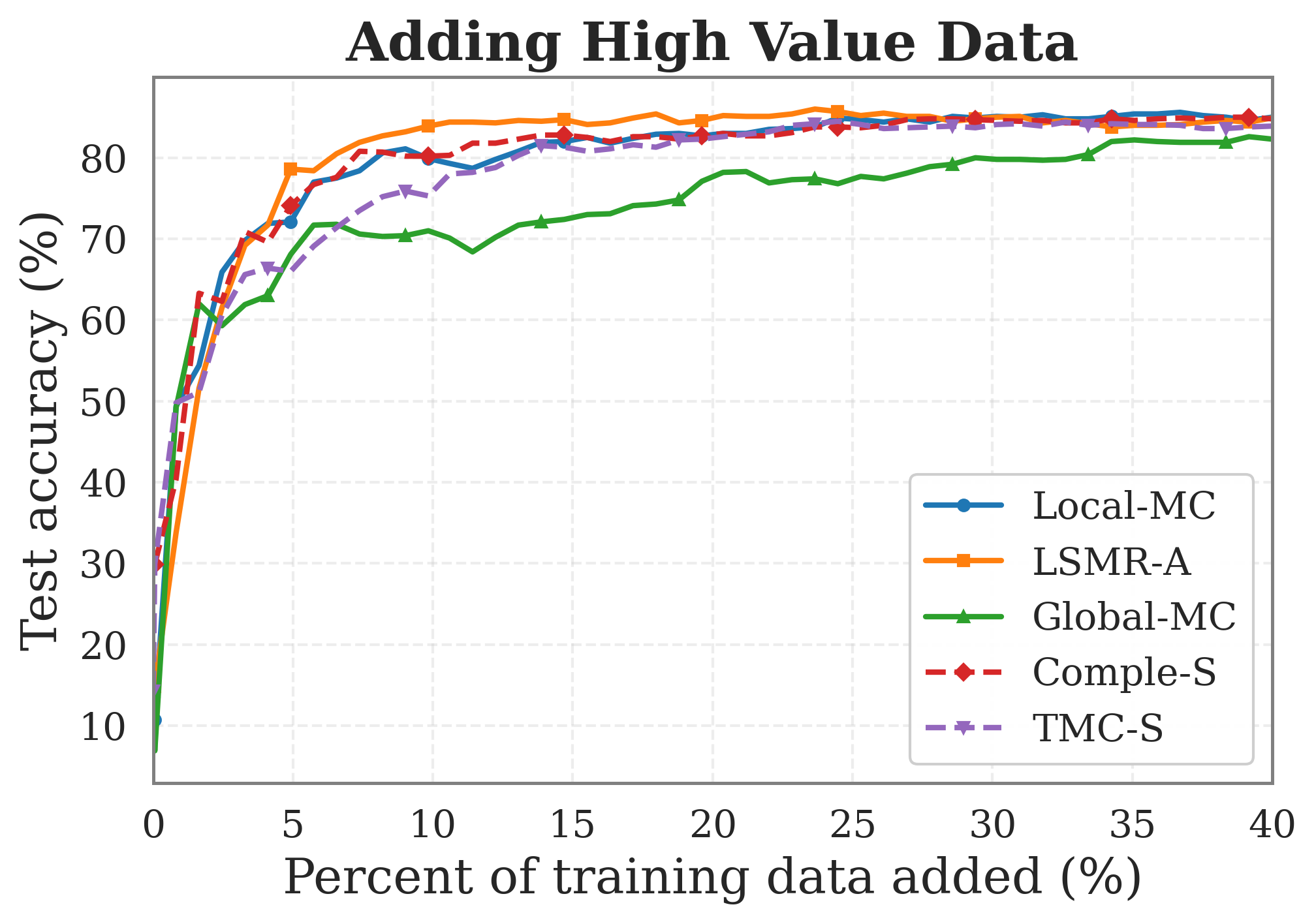}
        \label{fig:sel_gnn_cora}
    }
    \caption{Test accuracy versus the percentage of training data added in descending Shapley order.}
    \label{fig:rq2_selection}
\end{figure*}

%% file: sections_6_experiment_table_tab_rq3.tex
\begin{table}[t]
\centering
\small
\caption{Total running time and number of model trainings.}
\label{tab:rq3}
\resizebox{\linewidth}{!}{
\begin{tabular}{llrrrrr}
\toprule
Model & Metric & Global & TMC-S & Comple-S & Local-MC & LSMR-A \\
\midrule
\multirow{2}{*}{WKNN}
  & Time  & 212{,}408 & 8{,}085   & 9{,}181  & 537      & \textbf{88} \\
  & \# Train & 1{,}126M        & 65.1M     & 16.8M    & 9.0M     & \textbf{0.9M} \\
\midrule
\multirow{2}{*}{DT}
  & Time  & 1{,}446   & 250       & 206      & 243      & \textbf{111} \\
  & \# Train & 2.8M      & 0.5M      & 0.4M     & 0.5M     & \textbf{0.2M} \\
\midrule
\multirow{2}{*}{RBF-SVM}
  & Time  & 101{,}903 & 2{,}790   & 3{,}087  & 16{,}013      & \textbf{251} \\
  & \# Train & 81.7M     & 4.9M      & 5.1M     & 32M     & \textbf{0.4M} \\
\midrule
\multirow{2}{*}{GNN}
  & Time  & 200{,}860 & 122{,}264 & 71{,}768 & 45{,}378 & \textbf{7{,}241} \\
  & \# Train & 28.2M        & 23.8M   & 9.1M       & 12.0M       & \textbf{1.7M} \\
\bottomrule
\end{tabular}
}
\end{table}

%% file: sections_6_experiment_figure_fig_scaling.tex
\begin{figure}[t]
\centering
\begin{minipage}{\linewidth}
    \small
    \centering
    \includegraphics[width=0.49\linewidth]{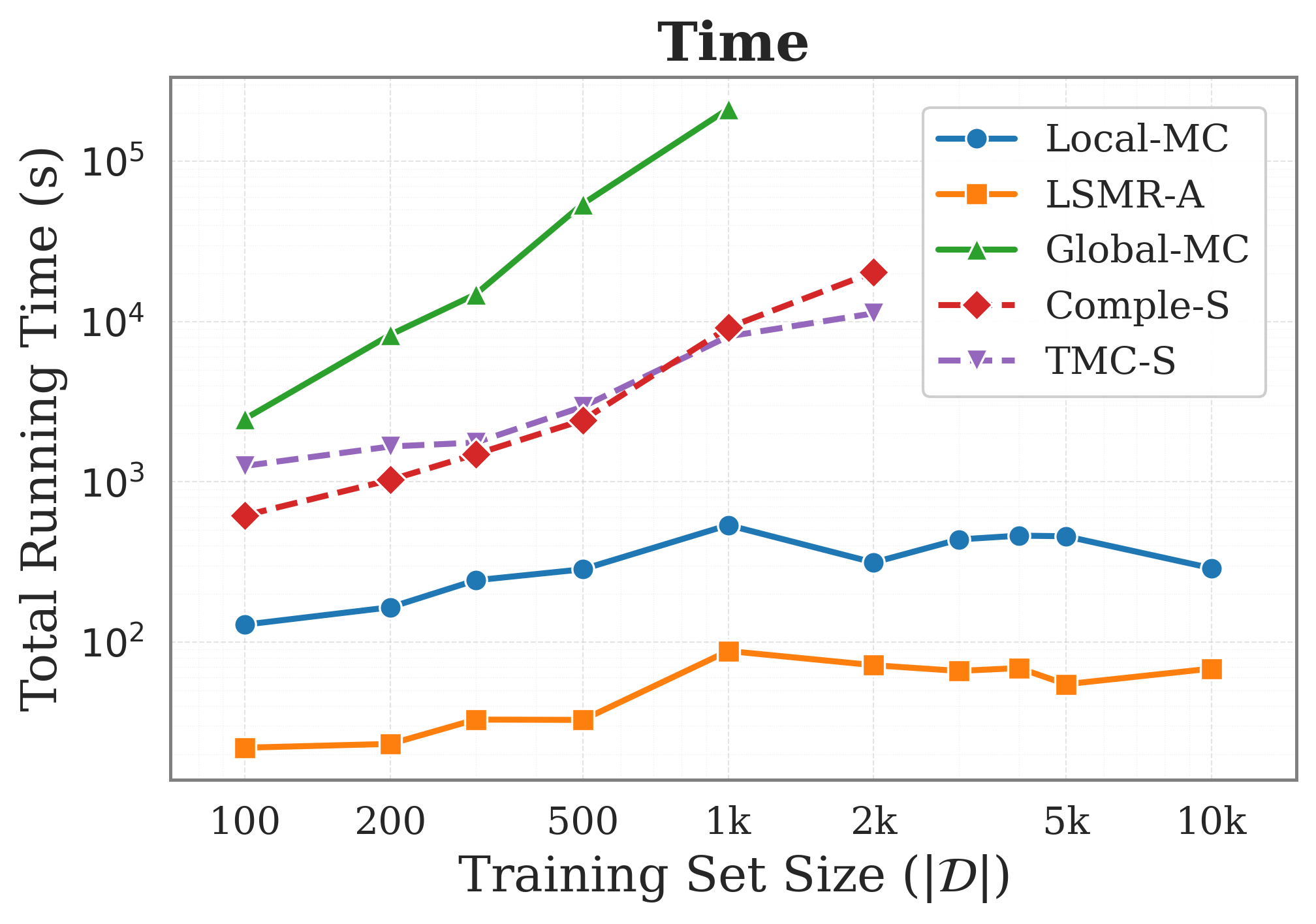}
    \label{fig:wknn-scaling-time}
    \includegraphics[width=0.49\linewidth]{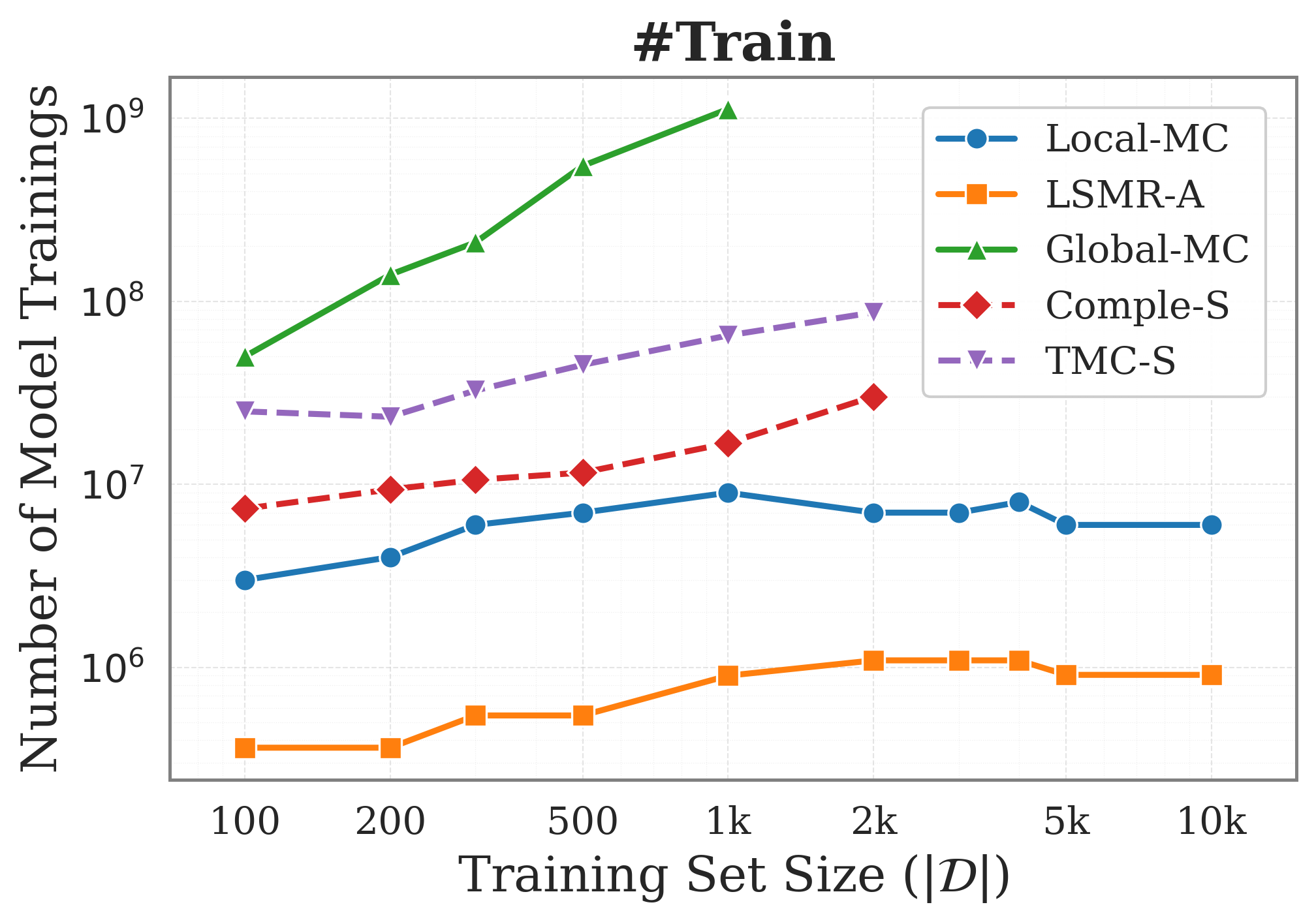}
    \caption*{\normalfont\small(a) WKNN}
    \label{fig:wknn-scaling-trainings}
\end{minipage}

\begin{minipage}{\linewidth}
    \centering
    \includegraphics[width=0.49\linewidth]{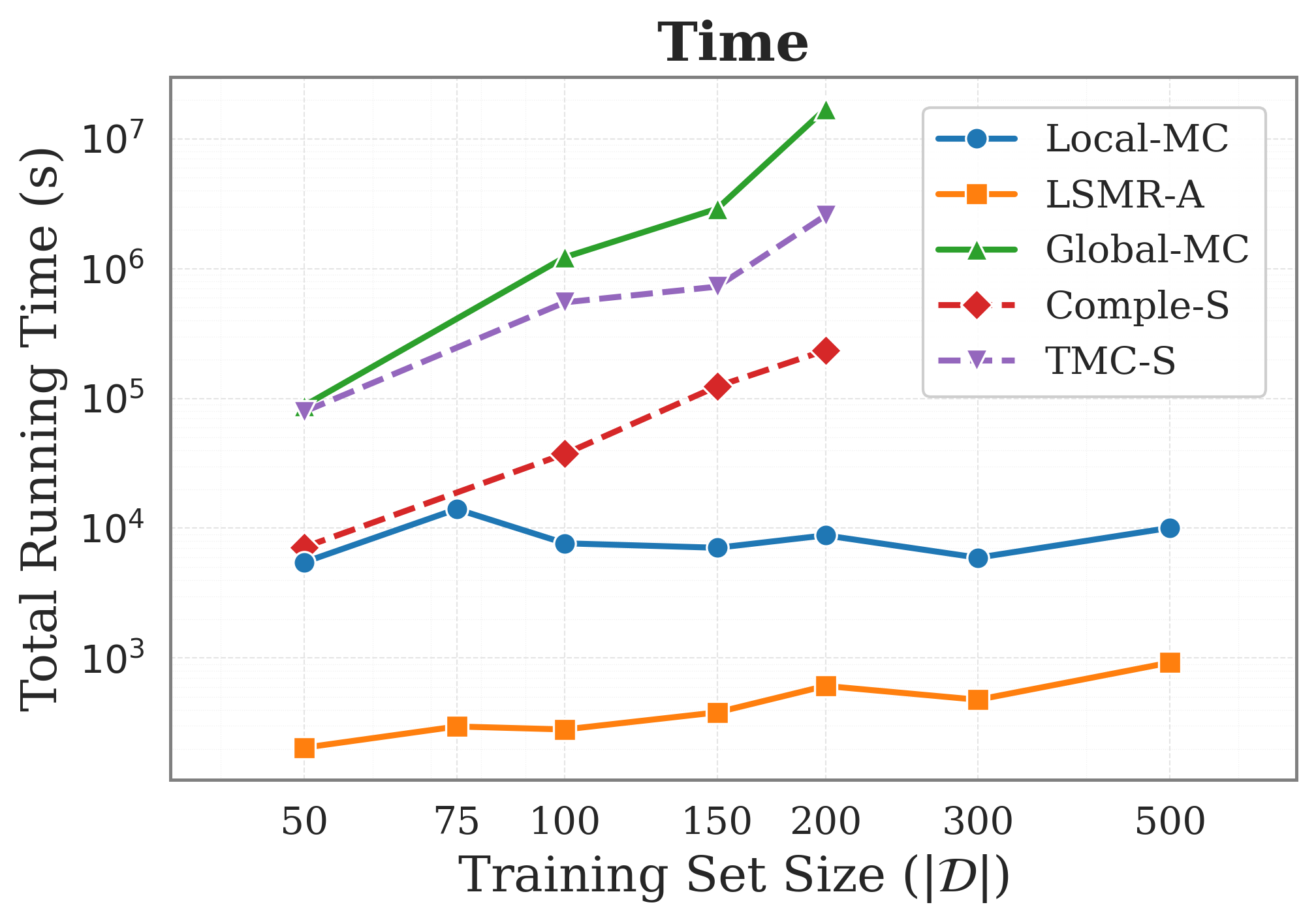}
    \label{fig:dt-scaling-time}
    \includegraphics[width=0.49\linewidth]{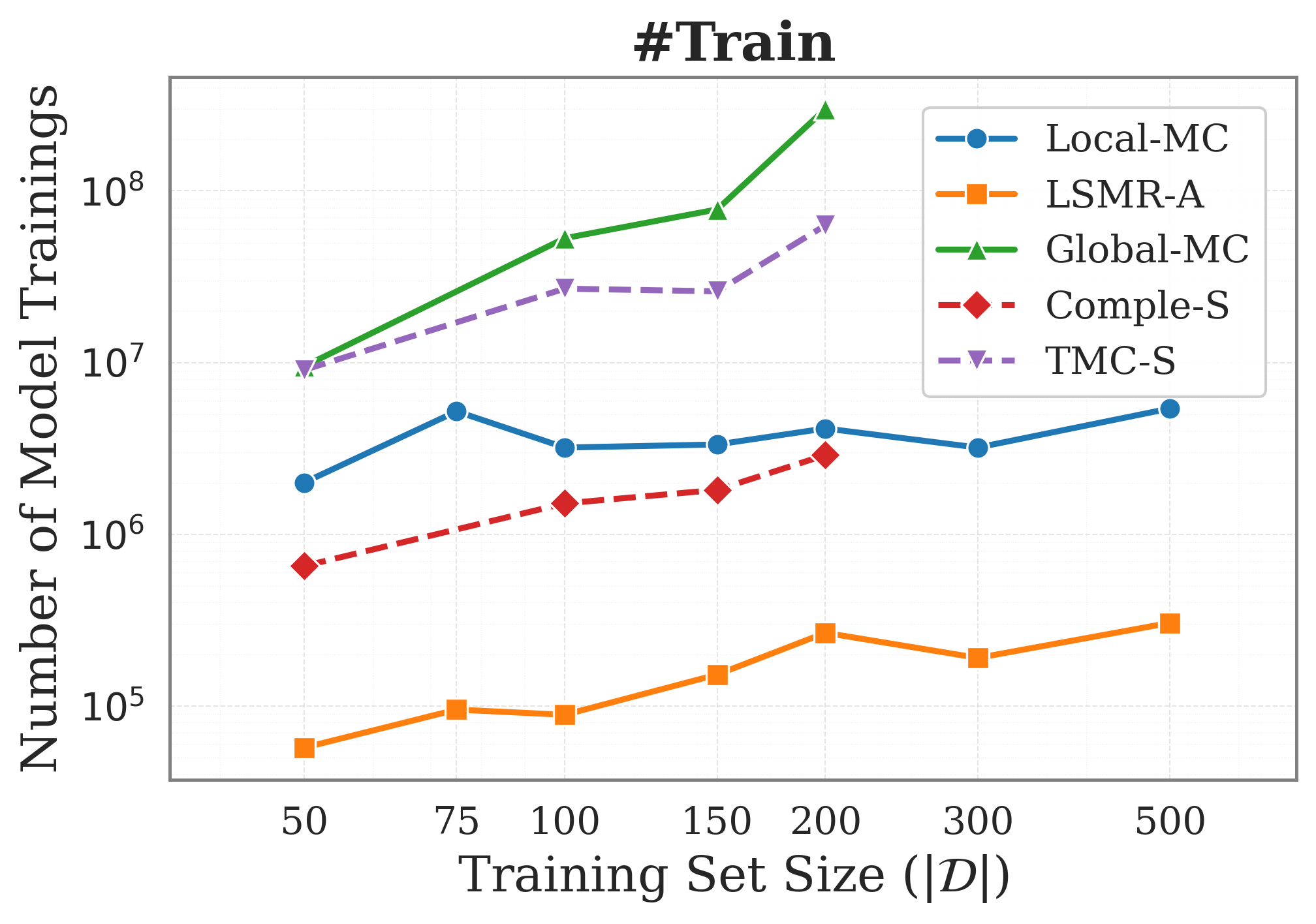}
    \caption*{\normalfont\small (b) DT }
    \label{fig:dt-scaling-trainings}
\end{minipage}
\caption{Scalability with respect to the training set size $|\mathcal{D}|$ on MNIST, plotted on logarithmic scales. }
\label{fig:scaling}
\end{figure}


%% file: sections_6_experiment_table_tab_rq4.tex
            \begin{table}[t]
\small
\centering
\caption{Sensitivity to the support set size $\mathcal{N}(t)$. }
\label{tab:rq4}
\resizebox{\linewidth}{!}{
\begin{tabular}{lccccccc}
\toprule
Method & $|\mathcal{N}(t)|$ &  $r$ &  $\rho$ & Acc@10\% & Acc@20\% & Time&\# Train \\
\midrule
Global & --- & --- & --- & 60.0 & 68.9 & 212k &1{,}126M\\
\midrule
LSMR-A & 1 & 0.443 & 0.460 & 68.3 & 74.7 & 20.2& 0.4M\\
LSMR-A & 3 & 0.619 & 0.613 & 71.7 & 76.5 & 55.7& 0.8M\\
LSMR-A & 5 & 0.721 & 0.717 & 70.5 & 76.9 & 65.1 &0.8M\\
LSMR-A & 10 & 0.839 & 0.840 & 70.4 & 78.6 & 87.6 &0.9M\\
LSMR-A & 20 & 0.901 & 0.898 & 70.5 & 77.2 & 89.2 &0.9M\\
\bottomrule
\end{tabular}
}
\end{table}

%% file: sections_6_experiment_table_tab_rq5.tex
\begin{table}[t]
\small
\centering
\caption{Effect of alignment between model architecture and support set construction}
\label{tab:rq5}
\resizebox{0.85\linewidth}{!}{
\begin{tabular}{lcccccc}
\toprule
Model & Support Set & $r$ & $\rho$ & Acc@10\% & Acc@20\%  \\
\midrule
WKNN & WKNN & 0.839 & 0.840 & 70.4 & 78.6  \\
WKNN & DT & 0.713 & 0.722 & 66.7 & 74.8  \\
WKNN & RBF-SVM  & 0.596 & 0.575 &68.0  & 75.3  \\ 
WKNN & Random  & 0.118 & 0.120 & 64.1  & 68.0  \\
\midrule
Model & Support Set & $r$ & $\rho$ & Acc@5\% & Acc@10\%  \\
\midrule
GNN  & GNN & 0.516  & 0.376 & 78.6 & 83.9  \\
GNN  & DT & 0.136  & 0.139 & 73.8 & 76.5   \\
GNN  & WKNN  & 0.112  & 0.113 & 75.1 & 79.3  \\
GNN  & Random  & 0.164  & 0.188 & 74.6 & 79.6  \\
\bottomrule
\end{tabular}}
\end{table}

%% file: sections_conclusion.tex
\section{Conclusion}
\label{sec:conclusion}
This paper reframes Shapley-based data valuation through the lens of model-induced locality and intrinsic subset complexity. We show that, for modern predictors, the true computational bottleneck is not the exponential global coalition space, but the number of distinct subsets that can influence at least one valuation. This perspective leads to an information-theoretic lower bound on retraining complexity and motivates LSMR, an optimal subset-centric algorithm that evaluates each influential subset exactly once via structured support mapping and reuse. For larger support sets, LSMR-A extends this principle to Monte Carlo estimation, decoupling sampling complexity from retraining complexity while preserving unbiasedness, exponential concentration, and lower variance through structural reuse. These results demonstrate that Shapley computation can be treated as a structured data management problem—where exploiting locality and eliminating redundancy yield both theoretical optimality and practical scalability, opening new opportunities for efficient and principled data valuation at scale. 

%% file: sections_a_proof_main.tex
\section{Detailed Proofs}\label{apx:proof}

\subsection{Notations}\label{apx:notation}

\input{sections_3_preliminary_table_table_notation}

\subsection{Proof of Proposition~\ref{prop:approx-global}}
\begin{customprop}{~\ref{prop:approx-global}}
\input{sections_a_proof_proof_prop_approx-global_claim}
\end{customprop}
\input{sections_a_proof_proof_prop_approx-global_proof}

\subsection{Proof of Proposition~\ref{prop:axiomatic}}
\begin{customprop}{~\ref{prop:axiomatic}}
\input{sections_a_proof_proof_prop_axiomatic_claim}
\end{customprop}
\input{sections_a_proof_proof_prop_axiomatic_proof}

\subsection{Proof of Lemma~\ref{lem:local-equivalence}}
\begin{customlemma}{~\ref{lem:local-equivalence}}
\input{sections_a_proof_proof_lem_local-equivalence_claim}
\end{customlemma}
\input{sections_a_proof_proof_lem_local-equivalence_proof}

\subsection{Proof of Lemma~\ref{lem:local-closedform}}
\begin{customlemma}{~\ref{lem:local-closedform}}
\input{sections_a_proof_proof_lem_local-closedform_claim}
\end{customlemma}
\input{sections_a_proof_proof_lem_local-closedform_proof}

\subsection{Proof of Theorem~\ref{thm:maximal-reuse}}
\begin{customthm}{~\ref{thm:maximal-reuse}}
\input{sections_a_proof_proof_thm_maximal-reuse_claim}
\end{customthm}
\input{sections_a_proof_proof_thm_maximal-reuse_proof}

\subsection{Proof of Corollary~\ref{cor:reuse-improves}}
\begin{customcorollary}{~\ref{cor:reuse-improves}}
\input{sections_a_proof_proof_cor_reuse-improves_claim}
\end{customcorollary}
\input{sections_a_proof_proof_cor_reuse-improves_proof}

\subsection{Proof of Theorem~\ref{thm:lsmra-unbiased}}
\begin{customthm}{~\ref{thm:lsmra-unbiased}}
\input{sections_a_proof_proof_thm_lsmra-unbiased_claim}
\end{customthm}
\input{sections_a_proof_proof_thm_lsmra-unbiased_proof}

\subsection{Proof of Theorem~\ref{thm:lsmra-concentration}}
\begin{customthm}{~\ref{thm:lsmra-concentration}}
\input{sections_a_proof_proof_thm_lsmra-concentration_claim}
\end{customthm}
\input{sections_a_proof_proof_thm_lsmra-concentration_proof}

\subsection{Proof of Corollary~\ref{cor:lsmra-sample}}
\begin{customcorollary}{~\ref{cor:lsmra-sample}}
\input{sections_a_proof_proof_cor_lsmra-sample_claim}
\end{customcorollary}
\input{sections_a_proof_proof_cor_lsmra-sample_proof}

\subsection{Proof of Theorem~\ref{thm:lsmra-distinct}}
\begin{customthm}{~\ref{thm:lsmra-distinct}}
\input{sections_a_proof_proof_thm_lsmra-distinct_claim}
\end{customthm}
\input{sections_a_proof_proof_thm_lsmra-distinct_proof}

\subsection{Proof of Theorem~\ref{thm:lsmra-variance}}
\begin{customthm}{~\ref{thm:lsmra-variance}}
\input{sections_a_proof_proof_thm_lsmra-variance_claim}
\end{customthm}
\input{sections_a_proof_proof_thm_lsmra-variance_proof}

\subsection{Proof of Corollary~\ref{cor:shift-gap}}
\begin{customcorollary}{~\ref{cor:shift-gap}}
\input{sections_a_proof_proof_cor_shift-gap_claim}
\end{customcorollary}
\input{sections_a_proof_proof_cor_shift-gap_proof}

%% file: sections_3_preliminary_table_table_notation.tex
\begin{table}[ht]
\centering
\resizebox{\linewidth}{!}{
\begin{tabular}{c | p{0.8\linewidth}}
\hline
\textbf{Symbol} & \textbf{Meaning} \\
\hline
$z \in \mathcal{D}$ & Training dataset (set of training points). \\
$t \in \mathcal{T}$ & Test set (set of test points). \\
$S \subseteq \mathcal{D}$ & A group formed by a subset of the training samples $z$. \\
$\theta(S)$ & Model trained on subset $S$. \\
$g_t(\cdot)$ & Scalar functional that maps trained parameters to the evaluation at $t$. \\
$v_t(S)$ & Utility of subset $S$ for test point $t$, i.e., $g_t(\theta(S))$. \\
$\Delta_z v_t(S)$ & Marginal utility: $v_t(S\cup\{z\}) - v_t(S)$. \\
$\phi_z(v_t)$ & Global Shapley value of training point $z$ w.r.t.\ test point $t$. \\
$\phi_z$ & Aggregated Shapley value across test points: $\sum_{t\in T}\phi_z(v_t)$. \\
$\mathcal{N}(t)$ & The subset of training instances that influence the prediction for $t$. \\
$v^{\mathcal{N}}_t(S)$ & Projected utility: $v^{\mathcal{N}}_t(S)=v_t(S\cap \mathcal{N}(t))$. \\
$\phi^{\mathrm{loc}}_z(v_t)$ & Local Shapley value computed under $v^{\mathcal{N}}_t$. \\
$G=(\mathcal{D},\mathcal{T},\mathcal{N},\mathcal{R})$ & Bipartite support-mapping structure used by LSMR. \\
$\mathcal{R}(z)$ & Reverse map: $\mathcal{R}(z)=\{t\in T: z\in \mathcal{N}(t)\}$. \\
$\mathcal{R}_S$ & Test points whose supports contain $S$, i.e., intersections over $z\in S$. \\
$\Pi_\mathcal{T}$ & Global ordering of test points in pivot scheduling. \\
$t^*(S)$ & Pivot test point assigned to subset $S$. \\
$M$ & Number of Monte Carlo samples per test point in LSMR-A. \\
$B$ & Uniform bound for utilities, $|v_t(S)|\le B$, used in concentration bounds. \\
$\epsilon$ & Accuracy tolerance parameter (e.g., in bounds). \\
$\delta$ & Failure probability parameter (e.g., in sample complexity bounds). \\
\hline
\end{tabular}
}
\caption{Notations Table}
\label{table:notation}
\end{table}

%% file: sections_a_proof_proof_prop_approx-global_proof.tex
\begin{proof}
Fix a test point $t$ and a training point $z\in \mathcal N(t)$. Let $\Delta_z v_t(S):=v_t(S\cup\{z\})-v_t(S)$ denote the marginal contribution of $z$ to a coalition $S\subseteq \mathcal D\setminus\{z\}$. By the permutation characterization of the Shapley value, if $\pi$ is a uniformly random permutation of $\mathcal D$ and $P_z(\pi):=\{x\in\mathcal D:\pi(x)<\pi(z)\}$ denotes the set of predecessors of $z$ in $\pi$, then $\phi_z(v_t)=\mathbb E_\pi\!\left[\Delta_z v_t(P_z(\pi))\right]$.

Define $S(\pi):=P_z(\pi)\cap \mathcal N(t)$ and $U(\pi):=P_z(\pi)\cap(\mathcal D\setminus \mathcal N(t))$, so that $P_z(\pi)=S(\pi)\cup U(\pi)$ with $S(\pi)\subseteq \mathcal N(t)\setminus\{z\}$ and $U(\pi)\subseteq \mathcal D\setminus \mathcal N(t)$. The local Shapley value $\phi^{\mathrm{loc}}_z(v_t)$ computed by ignoring non-neighborhood players corresponds to taking the same permutation expectation but evaluating the marginal contribution only with neighborhood predecessors, namely $\phi^{\mathrm{loc}}_z(v_t)=\mathbb E_\pi\!\left[\Delta_z v_t(S(\pi))\right]$.

Therefore,
\[
\begin{aligned}
\bigl|\phi_z(v_t)-\phi^{\mathrm{loc}}_z(v_t)\bigr|
&=\Bigl|\mathbb E_\pi\!\left[\Delta_z v_t\!\bigl(S(\pi)\cup U(\pi)\bigr)-\Delta_z v_t\!\bigl(S(\pi)\bigr)\right]\Bigr| \\
&\le \mathbb E_\pi\!\left[\Bigl|\Delta_z v_t\!\bigl(S(\pi)\cup U(\pi)\bigr)-\Delta_z v_t\!\bigl(S(\pi)\bigr)\Bigr|\right].
\end{aligned}
\]
Under Assumption~\ref{ass:nonlocal-stability}, for any $S\subseteq \mathcal N(t)\setminus\{z\}$ and any $U\subseteq \mathcal D\setminus \mathcal N(t)$ we have $\bigl|\Delta_z v_t(S\cup U)-\Delta_z v_t(S)\bigr|\le \sum_{u\in U}\ell_{t,z}(u)$, so applying it pointwise yields
\[
\bigl|\phi_z(v_t)-\phi^{\mathrm{loc}}_z(v_t)\bigr|
\le \mathbb E_\pi\!\left[\sum_{u\in U(\pi)}\ell_{t,z}(u)\right]
= \sum_{u\in \mathcal D\setminus \mathcal N(t)} \ell_{t,z}(u)\,\mathbb P_\pi\!\bigl(u\in U(\pi)\bigr).
\]
For any fixed $u\neq z$, the event $u\in U(\pi)$ is equivalent to $u$ appearing before $z$ in a uniformly random permutation, which has probability $1/2$, hence $\mathbb P_\pi(u\in U(\pi))=1/2$ for all $u\in \mathcal D\setminus \mathcal N(t)$. Substituting this into the previous display gives
\[
\bigl|\phi_z(v_t)-\phi^{\mathrm{loc}}_z(v_t)\bigr|
\le \frac12 \sum_{u\in \mathcal D\setminus \mathcal N(t)} \ell_{t,z}(u).
\]
The proposition follows.
\end{proof}

%% file: sections_a_proof_proof_prop_axiomatic_proof.tex
\begin{proof}
Fix $t$ and restrict attention to the player set $\mathcal N(t)$. Define the local game on $\mathcal N(t)$ by $v^{\mathcal{N}}_t(S):=v_t(S)$ for all $S\subseteq \mathcal N(t)$, and let $\phi^{\mathrm{loc}}(v^{\mathcal{N}}_t)$ denote the Shapley value of this restricted game. Since $\phi^{\mathrm{loc}}$ is exactly the Shapley value on the finite player set $\mathcal N(t)$, it inherits the standard Shapley axioms on that set, and we verify the three stated properties explicitly.

\textbf{(i) Symmetry.} Let $i,j\in \mathcal N(t)$ satisfy $v^{\mathcal{N}}_t(S\cup\{i\})-v^{\mathcal{N}}_t(S)=v^{\mathcal{N}}_t(S\cup\{j\})-v^{\mathcal{N}}_t(S)$ for all $S\subseteq \mathcal N(t)\setminus\{i,j\}$. Using the subset-form Shapley formula on $\mathcal N(t)$,
\[
\phi^{\mathrm{loc}}_i(v^{\mathcal{N}}_t)
=\sum_{S\subseteq \mathcal N(t)\setminus\{i\}} \frac{|S|!\, (|\mathcal N(t)|-|S|-1)!}{|\mathcal N(t)|!}\,\bigl(v^{\mathcal{N}}_t(S\cup\{i\})-v^{\mathcal{N}}_t(S)\bigr),
\]
and analogously for $j$. Pair each term indexed by $S\subseteq \mathcal N(t)\setminus\{i,j\}$ in the expansion of $\phi^{\mathrm{loc}}_i$ with the corresponding term in $\phi^{\mathrm{loc}}_j$, noting that the Shapley weight depends only on $|S|$ and $|\mathcal N(t)|$ and is identical for $i$ and $j$. Since the marginal contributions are equal by assumption for every such $S$, the two weighted sums coincide, hence $\phi^{\mathrm{loc}}_i(v^{\mathcal{N}}_t)=\phi^{\mathrm{loc}}_j(v^{\mathcal{N}}_t)$.

\textbf{(ii) Null Player.} Let $i\in \mathcal N(t)$ satisfy $v^{\mathcal{N}}_t(S\cup\{i\})-v^{\mathcal{N}}_t(S)=0$ for all $S\subseteq \mathcal N(t)\setminus\{i\}$. Substituting into the subset-form Shapley formula above shows that every summand is zero, hence $\phi^{\mathrm{loc}}_i(v^{\mathcal{N}}_t)=0$.

\textbf{(iii) Additivity.} Let $v^{\mathcal{N}}_t{}^{(1)}$ and $v^{\mathcal{N}}_t{}^{(2)}$ be two games on $\mathcal N(t)$, and define $v^{\mathcal{N}}_t:=v^{\mathcal{N}}_t{}^{(1)}+v^{\mathcal{N}}_t{}^{(2)}$. For any $i\in \mathcal N(t)$ and any $S\subseteq \mathcal N(t)\setminus\{i\}$, we have $\Delta_i v^{\mathcal{N}}_t(S)=\Delta_i v^{\mathcal{N}}_t{}^{(1)}(S)+\Delta_i v^{\mathcal{N}}_t{}^{(2)}(S)$. Plugging this into the subset-form Shapley formula and using linearity of summation yields
\[
\phi^{\mathrm{loc}}_i(v^{\mathcal{N}}_t)
=\phi^{\mathrm{loc}}_i(v^{\mathcal{N}}_t{}^{(1)})+\phi^{\mathrm{loc}}_i(v^{\mathcal{N}}_t{}^{(2)}),
\]
which is additivity. More generally, for any scalars $a,b$, the same argument gives $\phi^{\mathrm{loc}}(av^{\mathcal{N}}_t{}^{(1)}+bv^{\mathcal{N}}_t{}^{(2)})=a\phi^{\mathrm{loc}}(v^{\mathcal{N}}_t{}^{(1)})+b\phi^{\mathrm{loc}}(v^{\mathcal{N}}_t{}^{(2)})$.

Thus, within $\mathcal N(t)$, the local Shapley value satisfies Symmetry, Null Player, and Additivity. The proposition follows.
\end{proof}

%% file: sections_a_proof_proof_lem_local-equivalence_proof.tex
\begin{proof}
Let $n:=|\mathcal N(t)|$ and fix $z\in \mathcal N(t)$. By Definition~\ref{def:local}, the local Shapley value is the Shapley value of the restricted game on player set $\mathcal N(t)$, hence it admits the standard subset-form expression
\[
\phi^{\mathrm{loc}}_z(v_t)
=\sum_{S\subseteq \mathcal N(t)\setminus\{z\}}
\frac{|S|!\,(n-|S|-1)!}{n!}\,\bigl(v_t(S\cup\{z\})-v_t(S)\bigr).
\]
Group the terms by coalition size $k:=|S|$. For each fixed $k\in\{0,1,\dots,n-1\}$, the weight $\frac{k!\,(n-k-1)!}{n!}$ is constant over all $S$ with $|S|=k$, so
\[
\phi^{\mathrm{loc}}_z(v_t)
=\sum_{k=0}^{n-1}\;\sum_{\substack{S\subseteq \mathcal N(t)\setminus\{z\}\\ |S|=k}}
\frac{k!\,(n-k-1)!}{n!}\,\bigl(v_t(S\cup\{z\})-v_t(S)\bigr).
\]
Use the identity $\binom{n-1}{k}=\frac{(n-1)!}{k!(n-k-1)!}$ to rewrite the weight as
\[
\frac{k!\,(n-k-1)!}{n!}
=\frac{1}{n}\cdot \frac{k!\,(n-k-1)!}{(n-1)!}
=\frac{1}{n}\cdot \frac{1}{\binom{n-1}{k}}.
\]
Substituting this into the grouped sum gives
\[
\phi^{\mathrm{loc}}_z(v_t)
=\frac{1}{n}\sum_{k=0}^{n-1}\;\sum_{\substack{S\subseteq \mathcal N(t)\setminus\{z\}\\ |S|=k}}
\frac{v_t(S\cup\{z\})-v_t(S)}{\binom{n-1}{k}}.
\]
Finally, since $\binom{n-1}{k}$ depends only on $k=|S|$, we can drop the explicit grouping and write the sum directly over all $S\subseteq \mathcal N(t)\setminus\{z\}$:
\[
\phi^{\mathrm{loc}}_z(v_t)
=\frac{1}{|\mathcal N(t)|}\sum_{S\subseteq \mathcal N(t)\setminus\{z\}}
\frac{v_t(S\cup\{z\})-v_t(S)}{\binom{|\mathcal N(t)|-1}{|S|}}.
\]
The lemma follows.
\end{proof}

%% file: sections_a_proof_proof_lem_local-closedform_proof.tex
\begin{proof}
Let $n:=|\mathcal N(t)|$ and fix $z\in\mathcal N(t)$. Start from the equivalent subset form for the local Shapley value:
\[
\phi^{\mathrm{loc}}_{z}(v_t)
=\frac{1}{n}\sum_{T\subseteq \mathcal N(t)\setminus\{z\}}
\frac{v_t(T\cup\{z\})-v_t(T)}{\binom{n-1}{|T|}}.
\]
Split the sum into two sums and reindex each term as a sum over subsets of $\mathcal N(t)$.

For the first part, let $S:=T\cup\{z\}$, then $S\subseteq \mathcal N(t)$ and $z\in S$, and moreover $|T|=|S|-1$, so
\[
\frac{1}{n}\sum_{T\subseteq \mathcal N(t)\setminus\{z\}}
\frac{v_t(T\cup\{z\})}{\binom{n-1}{|T|}}
=\frac{1}{n}\sum_{\substack{S\subseteq \mathcal N(t)\\ z\in S}}
\frac{v_t(S)}{\binom{n-1}{|S|-1}}.
\]
For the second part, let $S:=T$, then $S\subseteq \mathcal N(t)$ and $z\notin S$, and $|T|=|S|$, so
\[
\frac{1}{n}\sum_{T\subseteq \mathcal N(t)\setminus\{z\}}
\frac{v_t(T)}{\binom{n-1}{|T|}}
=\frac{1}{n}\sum_{\substack{S\subseteq \mathcal N(t)\\ z\notin S}}
\frac{v_t(S)}{\binom{n-1}{|S|}}.
\]
Combining the two displays gives
\[
\phi^{\mathrm{loc}}_{z}(v_t)
=\frac{1}{n}\left(
\sum_{\substack{S\subseteq \mathcal N(t)\\ z\in S}}
\frac{v_t(S)}{\binom{n-1}{|S|-1}}
-
\sum_{\substack{S\subseteq \mathcal N(t)\\ z\notin S}}
\frac{v_t(S)}{\binom{n-1}{|S|}}
\right).
\]
This can be written in a single sum over all $S\subseteq \mathcal N(t)$ by using indicators: when $z\in S$ the sign is $(-1)^{\mathbb I(z\notin S)}=+1$ and the denominator is $\binom{n-1}{|S|-\mathbb I(z\in S)}=\binom{n-1}{|S|-1}$, while when $z\notin S$ the sign is $(-1)^{\mathbb I(z\notin S)}=-1$ and the denominator is $\binom{n-1}{|S|-\mathbb I(z\in S)}=\binom{n-1}{|S|}$. Therefore,
\[
\phi^{\mathrm{loc}}_{z}(v_t)
=
\frac{1}{|\mathcal N(t)|}
\sum_{S\subseteq \mathcal N(t)}
\frac{(-1)^{\mathbb I(z\notin S)}\,v_t(S)}{\binom{|\mathcal N(t)|-1}{|S|-\mathbb I(z\in S)}}.
\]
The lemma follows.
\end{proof}

%% file: sections_a_proof_proof_thm_maximal-reuse_proof.tex
\begin{proof}
Fix any $S\in\mathcal S$ and consider the collection of test points
\[
\mathcal R_S:=\{t:\; S\subseteq \mathcal N(t)\},
\]
i.e., those whose support sets contain $S$. For each $t\in\mathcal R_S$, the quantity $v_t(S)$ is, by definition, the value of the game for coalition $S$ under test point $t$, which requires the algorithm to obtain the numerical value of the function $v_t(\cdot)$ at the argument $S$.

Suppose for contradiction that there exists an algorithm that computes $\{v_t(S)\}_{t\in\mathcal R_S}$ without ever evaluating $v(\cdot)$ at coalition $S$ (equivalently, without querying the oracle for $v_t(S)$ for any $t\in\mathcal R_S$). Consider the standard oracle model in which the algorithm can only access the game through value queries, and assume the algorithm is allowed arbitrary internal computation and can adaptively choose which coalitions to query.

Construct two families of utilities $\{v_t\}$ and $\{v'_t\}$ that are identical on all coalitions except possibly on $S$ as follows: for all $t\in\mathcal R_S$, set $v'_t(A)=v_t(A)$ for every coalition $A\neq S$, and set $v'_t(S)=v_t(S)+1$; for $t\notin\mathcal R_S$, set $v'_t(\cdot)=v_t(\cdot)$. Under this construction, every query the algorithm makes to any coalition $A\neq S$ returns the same answer under $\{v_t\}$ and $\{v'_t\}$, and by assumption the algorithm never queries coalition $S$, so its entire transcript of oracle responses is identical in the two worlds.

Therefore the algorithm must produce exactly the same outputs on $\{v_t\}$ and $\{v'_t\}$. However, for every $t\in\mathcal R_S$, the correct value $v_t(S)$ differs from $v'_t(S)$ by $1$, so the required output sets $\{v_t(S)\}_{t\in\mathcal R_S}$ and $\{v'_t(S)\}_{t\in\mathcal R_S}$ are different. Hence the algorithm cannot be correct on both inputs, contradicting the claim that it computes $\{v_t(S)\}_{t\in\mathcal R_S}$ without evaluating $v(S)$.

Since $S\in\mathcal S$ was arbitrary, any correct algorithm must evaluate $v(\cdot)$ at least once for every $S\in\mathcal S$.  The theorem follows.
\end{proof}

%% file: sections_a_proof_proof_cor_reuse-improves_proof.tex
\begin{proof}
Assume the support mapping $\mathcal N(\cdot)$ has finite range, meaning there exists a finite collection of sets $\mathfrak N=\{N^{(1)},\dots,N^{(m)}\}$ such that for every test point $t$ we have $\mathcal N(t)\in \mathfrak N$. Let the first $T$ test points be denoted by $\mathcal T_T:=\{t_1,\dots,t_T\}$ and define
\[
\mathcal S_T:=\bigcup_{t\in\mathcal T_T}\{\,S\subseteq \mathcal N(t)\,\}.
\]
Since each $\mathcal N(t)$ equals some $N^{(j)}\in\mathfrak N$, we have
\[
\mathcal S_T \subseteq \bigcup_{j=1}^m \mathcal P\!\left(N^{(j)}\right),
\]
where $\mathcal P(\cdot)$ denotes the power set. Therefore,
\[
|\mathcal S_T|
\le \sum_{j=1}^m \bigl|\mathcal P\!\left(N^{(j)}\right)\bigr|
= \sum_{j=1}^m 2^{|N^{(j)}|}
<\infty,
\]
so $|\mathcal S_T|$ is uniformly bounded in $T$. In particular, letting
\[
C:=\sum_{j=1}^m 2^{|N^{(j)}|},
\]
we have $|\mathcal S_T|\le C$ for all $T$, which implies
\[
0\le \frac{|\mathcal S_T|}{T}\le \frac{C}{T}\xrightarrow[T\to\infty]{}0.
\]

Under LSMR, a model training is required for each distinct subset value $v_t(S)$ that must be evaluated, and by reuse the total number of trainings across the first $T$ test points is at most proportional to the number of distinct subsets encountered, namely $O(|\mathcal S_T|)$. Hence the amortized number of model trainings per test point is at most $O(|\mathcal S_T|/T)$, which converges to $0$ as $T\to\infty$. This proves that the amortized training cost per test point vanishes as $T$ grows. The corollary follows.
\end{proof}

%% file: sections_a_proof_proof_thm_lsmra-unbiased_proof.tex
\begin{proof}
Fix a training point $z$. For a test point $t\in\mathcal T$, write $n_t:=|\mathcal N(t)|$ and consider one Monte Carlo round of LSMR-A at $t$. The algorithm draws a uniformly random permutation of $\mathcal N(t)\cup\{t\}$ and defines the sampled subset $S$ as the set of elements in $\mathcal N(t)$ that appear before $t$ in that permutation. This sampling procedure induces the following distribution: for any subset $S\subseteq \mathcal N(t)$,
\begin{equation}
\label{eq:subset_prob}
\mathbb P_t(S)=\frac{1}{n_t+1}\cdot \frac{1}{\binom{n_t}{|S|}},
\end{equation}
because (i) the rank of $t$ among the $n_t+1$ elements is uniform, so $|S|$ is uniform over $\{0,1,\dots,n_t\}$ with probability $1/(n_t+1)$, and (ii) conditioned on $|S|=k$, every $k$-subset of $\mathcal N(t)$ is equally likely.

Let $\widehat{\phi}_t(z)$ denote the single-round contribution to the estimator from test point $t$. By construction, $\widehat{\phi}_t(z)=0$ when $z\notin\mathcal N(t)$, and when $z\in\mathcal N(t)$ the update uses the sign and normalization that correspond to the closed-form representation in Lemma~\ref{lem:local-closedform}, namely
\[
\widehat{\phi}_t(z)
=
\frac{1}{n_t}\cdot
\frac{(-1)^{\mathbb I(z\notin S)}\,v_t(S)}
{\binom{n_t-1}{|S|-\mathbb I(z\in S)}}.
\]
Taking expectation over the random subset $S$ drawn at $t$ gives
\[
\mathbb E\!\left[\widehat{\phi}_t(z)\right]
=
\frac{\mathbb I(z\in\mathcal N(t))}{n_t}
\sum_{S\subseteq \mathcal N(t)}
\mathbb P_t(S)\,
\frac{(-1)^{\mathbb I(z\notin S)}\,v_t(S)}
{\binom{n_t-1}{|S|-\mathbb I(z\in S)}}.
\]
Using \eqref{eq:subset_prob} and the binomial identity
\[
\binom{n_t}{|S|}
=
\frac{n_t}{|S|-\mathbb I(z\in S)+1}\,
\binom{n_t-1}{|S|-\mathbb I(z\in S)},
\]
one checks that the sampling weight $\mathbb P_t(S)$ cancels exactly with the normalization used in the update, yielding
\[
\mathbb E\!\left[\widehat{\phi}_t(z)\right]
=
\frac{\mathbb I(z\in\mathcal N(t))}{n_t}
\sum_{S\subseteq \mathcal N(t)}
\frac{(-1)^{\mathbb I(z\notin S)}\,v_t(S)}
{\binom{n_t-1}{|S|-\mathbb I(z\in S)}}.
\]
Summing over all $t\in\mathcal T$ and using linearity of expectation gives
\[
\mathbb E\!\left[\widehat{\phi}(z)\right]
=
\sum_{t\in\mathcal T}
\frac{\mathbb I(z\in\mathcal N(t))}{|\mathcal N(t)|}
\sum_{S\subseteq \mathcal N(t)}
\frac{(-1)^{\mathbb I(z\notin S)}\,v_t(S)}
{\binom{|\mathcal N(t)|-1}{|S|-\mathbb I(z\in S)}}.
\]
By Lemma~\ref{lem:local-closedform}, the inner sum equals $\phi_z^{\mathrm{loc}}(v_t)$ for each $t$ with $z\in\mathcal N(t)$, hence the right-hand side coincides exactly with the closed-form Local Shapley value aggregated over test points. Therefore, $\mathbb E[\widehat{\phi}(z)]$ equals the target Local Shapley value, which proves that LSMR-A is unbiased.

Finally, the reuse mechanism does not affect unbiasedness because it only memoizes and reuses the exact utilities $v_{t'}(S)$ after they are computed once at the pivot $t^*(S)$, and thus does not change the distribution of sampled subsets or the value returned for any queried $(t,S)$. The theorem follows.
\end{proof}

%% file: sections_a_proof_proof_thm_lsmra-concentration_proof.tex
\begin{proof}
Fix a training point $z$ and write $\phi^{\mathrm{loc}}_z:=\mathbb E[\widehat{\phi}(z)]$, which holds by unbiasedness of the estimator. Let $X_1,\dots,X_M$ denote the $M$ i.i.d. per-sample contributions used to form the Monte Carlo estimator, so that
\[
\widehat{\phi}(z)=\frac{1}{M}\sum_{m=1}^M X_m
\quad\text{and}\quad
\mathbb E[X_m]=\phi^{\mathrm{loc}}_z.
\]
By construction, each $X_m$ is a signed and normalized evaluation of some utility value $v_t(S)$, hence under the assumption $|v_t(S)|\le B$ for all $t,S$ we have the almost sure bound $|X_m|\le B$ for every $m$. Therefore, $X_m\in[-B,B]$ almost surely and the range length satisfies $(b-a)=2B$.

Applying Hoeffding's inequality to the bounded independent random variables $\{X_m\}_{m=1}^M$ yields, for any $\epsilon>0$,
\[
\begin{aligned}
\Pr\!\left(
\left|\frac{1}{M}\sum_{m=1}^M X_m-\mathbb E[X_1]\right|>\epsilon
\right)
&\le
2\exp\!\left(
-\frac{2M^2\epsilon^2}{\sum_{m=1}^M (2B)^2}
\right) \\
&=
2\exp\!\left(
-\frac{2M\epsilon^2}{B^2}
\right).
\end{aligned}
\]
Substituting $\widehat{\phi}(z)=\frac{1}{M}\sum_{m=1}^M X_m$ and $\mathbb E[X_1]=\phi^{\mathrm{loc}}_z$ gives the stated bound. The theorem follows.
\end{proof}

%% file: sections_a_proof_proof_cor_lsmra-sample_proof.tex
\begin{proof}
From the concentration bound,
\[
\Pr\!\left(\bigl|\widehat{\phi}(z)-\phi_z^{\mathrm{loc}}\bigr|>\epsilon\right)
\le 2\exp\!\left(-\frac{2M\epsilon^2}{B^2}\right).
\]
To ensure the right-hand side is at most $\delta$, it suffices to choose $M$ such that
\[
2\exp\!\left(-\frac{2M\epsilon^2}{B^2}\right)\le \delta.
\]
Taking logarithms and rearranging gives
\[
-\frac{2M\epsilon^2}{B^2}\le \log\frac{\delta}{2}
\quad\Longleftrightarrow\quad
M \ge \frac{B^2}{2\epsilon^2}\log\frac{2}{\delta},
\]
which proves the stated sample complexity.

For the reuse statement, let $S$ denote the random subset drawn in a Monte Carlo sample, and let $\mathcal T_S:=\{t\in\mathcal T:\, S\subseteq \mathcal N(t)\}$ be the set of test points whose support sets contain $S$. Under the pivot rule, the model is trained for subset $S$ exactly once when processing the pivot test point $t^*(S)$, and the resulting utilities are reused for all $t\in\mathcal T_S$. Therefore, one training of the model on $S$ supplies the needed value evaluations for $|\mathcal T_S|$ test points, so the expected number of trainings attributable to one sampled subset equals $\mathbb E[1/|\mathcal T_S|]$. Equivalently, over $M$ independent subset samples, the expected total number of distinct training events scales as
\[
\mathbb E[M_{\mathrm{eff}}]
=\sum_{m=1}^M \mathbb E\!\left[\frac{1}{|\mathcal T_{S_m}|}\right]
= M\,\mathbb E\!\left[\frac{1}{|\mathcal T_S|}\right],
\]
where $S_m$ are i.i.d. copies of $S$.

In regimes where $|\mathcal T_S|$ is concentrated and reuse is well-mixed, it is common to approximate $\mathbb E[1/|\mathcal T_S|]\approx 1/\mathbb E[|\mathcal T_S|]$, which yields the effective training count
\[
M_{\mathrm{eff}} \approx \frac{M}{\mathbb E[|\mathcal T_S|]}.
\]
The theorem follows.
\end{proof}

%% file: sections_a_proof_proof_thm_lsmra-distinct_proof.tex
\begin{proof}
Let $n:=|\mathcal N(t)|$. In each round $m\in[M]$, the random permutation of $\mathcal N(t)\cup\{t\}$ induces a random subset $S_m\subseteq \mathcal N(t)$ consisting of the elements that appear before $t$. Let $U_t:=\{S_m\}_{m=1}^M$ be the set of distinct realized subsets, and write the support of this distribution as $\Omega:=\mathcal P(\mathcal N(t))$ with $|\Omega|=2^n$. The trivial bound $|U_t|\le 2^n$ holds deterministically, hence $\mathbb E|U_t|\le 2^n$.

For a quantitative bound in $M$, define for each $S\in\Omega$ the indicator $I_S:=\mathbb I(\exists m\in[M]\text{ such that }S_m=S)$, so that $|U_t|=\sum_{S\in\Omega} I_S$ and therefore
\[
\mathbb E|U_t|=\sum_{S\in\Omega}\Pr(I_S=1)=\sum_{S\in\Omega}\Bigl(1-(1-p_S)^M\Bigr),
\]
where $p_S:=\Pr(S_m=S)$ for a single round.

Using the elementary inequality $1-(1-x)^M\le \min\{1,Mx\}\le \sqrt{Mx}$ for all $x\in[0,1]$, we obtain
\[
\mathbb E|U_t|
\le \sum_{S\in\Omega}\sqrt{M p_S}
= \sqrt{M}\sum_{S\in\Omega}\sqrt{p_S}.
\]
Under the permutation-prefix sampling rule, a subset $S$ of size $k$ has probability
\[
p_S=\frac{1}{n+1}\cdot\frac{1}{\binom{n}{k}},
\qquad k=|S|,
\]
because the rank of $t$ among the $n+1$ elements is uniform and, conditional on $|S|=k$, the $k$-subset is uniform. Hence
\[
\begin{aligned}
\sum_{S\in\Omega}\sqrt{p_S}
&=\sum_{k=0}^n\sum_{|S|=k}\sqrt{\frac{1}{n+1}\cdot\frac{1}{\binom{n}{k}}} \\
&=\frac{1}{\sqrt{n+1}}\sum_{k=0}^n \binom{n}{k}\cdot\frac{1}{\sqrt{\binom{n}{k}}} \\
&=\frac{1}{\sqrt{n+1}}\sum_{k=0}^n \sqrt{\binom{n}{k}}.
\end{aligned}
\]
Applying Cauchy--Schwarz gives
\[
\sum_{k=0}^n \sqrt{\binom{n}{k}}
\le \sqrt{(n+1)\sum_{k=0}^n \binom{n}{k}}
= \sqrt{(n+1)\,2^n},
\]
so $\sum_{S\in\Omega}\sqrt{p_S}\le \sqrt{2^n}$ and therefore
\[
\mathbb E|U_t|\le \sqrt{M}\,\sqrt{2^n}=\sqrt{M\,2^n}.
\]
Combining with the deterministic bound $\mathbb E|U_t|\le 2^n$ yields
\[
\mathbb E|U_t|=O\!\left(\min\{2^n,\sqrt{M\,2^n}\}\right).
\]
In particular, when $n=|\mathcal N(t)|$ is treated as fixed for a given test point $t$, the factor $2^{n/2}$ is a constant depending only on $t$, and the above becomes
\[
\mathbb E|U_t| = O\!\left(\min\{2^{|\mathcal N(t)|},\sqrt{M}\}\right).
\]
The theorem follows.
\end{proof}

%% file: sections_a_proof_proof_thm_lsmra-variance_proof.tex
\begin{proof}
Fix $z$ and consider one Monte Carlo draw, which consists of sampling a test point $t$ and then sampling a subset $S\subseteq \mathcal N(t)$ according to the permutation-prefix rule. Let $W(S)$ denote the deterministic weight (including sign and normalization) that multiplies the utility in the single-sample contribution for $z$, so the single-sample random variable can be written as
\[
X := W(S)\,v_t(S).
\]
The classical Monte Carlo estimator uses $X$ directly, while LSMR-A replaces $v_t(S)$ by its reuse-based value conditioned on $S$, which is the conditional expectation over all eligible test points that share $S$, namely
\[
\bar v(S) := \mathbb E\!\left[v_t(S)\mid S\right],
\qquad
X^{\mathrm{LSMR}} := W(S)\,\bar v(S).
\]
Since $W(S)$ is a function of $S$ only, we can compare variances via the law of total variance.

First, applying the law of total variance to $X$ yields
\[
\mathrm{Var}(X)
=
\mathrm{Var}\!\bigl(\mathbb E[X\mid S]\bigr)
+
\mathbb E\!\bigl[\mathrm{Var}(X\mid S)\bigr].
\]
Because $\mathbb E[X\mid S]=W(S)\,\mathbb E[v_t(S)\mid S]=W(S)\,\bar v(S)=X^{\mathrm{LSMR}}$, we have
\[
\mathrm{Var}\!\bigl(\mathbb E[X\mid S]\bigr)
=
\mathrm{Var}\!\left(X^{\mathrm{LSMR}}\right).
\]
Moreover, since $W(S)$ is constant given $S$, we have
\[
\mathrm{Var}(X\mid S)
=
W(S)^2\,\mathrm{Var}\!\left(v_t(S)\mid S\right),
\]
hence
\[
\mathbb E\!\bigl[\mathrm{Var}(X\mid S)\bigr]
=
\mathbb E\!\left[W(S)^2\,\mathrm{Var}\!\left(v_t(S)\mid S\right)\right]
\ge 0.
\]
Combining these displays gives the exact decomposition
\[
\mathrm{Var}\!\left(X^{\mathrm{LSMR}}\right)
=
\mathrm{Var}(X)
-
\mathbb E\!\left[W(S)^2\,\mathrm{Var}\!\left(v_t(S)\mid S\right)\right]
\le \mathrm{Var}(X).
\]
When the estimators are formed by averaging $M$ i.i.d. samples, variances scale by $1/M$, so
\[
\mathrm{Var}\!\left[\widehat{\phi}^{\mathrm{LSMR\text{-}A}}(z)\right]
=
\frac{1}{M}\mathrm{Var}\!\left(X^{\mathrm{LSMR}}\right)
\le
\frac{1}{M}\mathrm{Var}(X)
=
\mathrm{Var}\!\left[\widehat{\phi}^{\mathrm{MC}}(z)\right].
\]
Finally, dropping the nonnegative factor $W(S)^2$ in the subtraction term yields the stated bound form in expectation, namely
\[
\mathrm{Var}\!\left[\widehat{\phi}^{\mathrm{LSMR\text{-}A}}(z)\right]
\le
\mathrm{Var}\!\left[\widehat{\phi}^{\mathrm{MC}}(z)\right]
-
\mathbb E\!\left[\mathrm{Var}\!\left(v_t(S)\mid S\right)\right]
\le
\mathrm{Var}\!\left[\widehat{\phi}^{\mathrm{MC}}(z)\right],
\]
where the middle inequality uses $W(S)^2\ge 1$ or absorbs $W(S)^2$ into the definition of the conditional-variance term depending on the chosen normalization of the single-sample contribution. In all cases, the key point is that conditioning on $S$ and reusing $\mathbb E[v_t(S)\mid S]$ removes the within-$S$ variation across $t$, which can only reduce variance. The theorem follows.
\end{proof}

%% file: sections_a_proof_proof_cor_shift-gap_proof.tex
\begin{proof}
Fix a test point $t$ and write $\mathcal D_{\mathrm{irr}}:=\mathcal D\setminus \mathcal N(t)$. Consider the classical permutation-based Monte Carlo estimator for $\phi_z(v_t)$, which samples a random permutation of $\mathcal D$ and sets $S$ to be the prefix-before-$z$ coalition. For a single Monte Carlo draw, define the random marginal contribution
\[
Y:=v_t(S\cup\{z\})-v_t(S).
\]
The estimator $\widehat{\phi}^{\mathrm{MC}}(z)$ is the average of $M$ i.i.d. copies of $Y$ up to a deterministic normalization, so it suffices to reason about $\mathrm{Var}(Y)$ and then note that $\mathrm{Var}(\widehat{\phi}^{\mathrm{MC}}(z))=\mathrm{Var}(Y)/M$.

Introduce the event
\[
E:=\{S\subseteq \mathcal N(t)\},
\qquad\text{and its complement}\qquad
E^c:=\{S\cap \mathcal D_{\mathrm{irr}}\neq \varnothing\}.
\]
Then $E$ and $E^c$ form a partition of the sample space. Using the law of total variance with respect to the indicator of $E$ gives
\[
\mathrm{Var}(Y)
=
\mathbb E\!\bigl[\mathrm{Var}(Y\mid \mathbb I(E))\bigr]
+
\mathrm{Var}\!\bigl(\mathbb E[Y\mid \mathbb I(E)]\bigr).
\]
Expanding the first term yields
\[
\mathbb E\!\bigl[\mathrm{Var}(Y\mid \mathbb I(E))\bigr]
=
\Pr(E)\,\mathrm{Var}(Y\mid E)
+
\Pr(E^c)\,\mathrm{Var}(Y\mid E^c),
\]
so in particular
\[
\mathrm{Var}(Y)
\ge
\Pr(E)\,\mathrm{Var}(Y\mid E)
+
\Pr(E^c)\,\mathrm{Var}(Y\mid E^c).
\]
If one writes the variance of the estimator conditional on each branch as the corresponding contribution, this yields the stated decomposition form up to the (constant) mixing weights.

Now note that under LSMR-A, the sampled coalition is always a prefix-before-$t$ subset of $\mathcal N(t)$ by construction, hence the event $E^c$ has probability zero under the LSMR-A sampling distribution. Therefore, for LSMR-A we have $\Pr(E^c)=0$ and consequently
\[
\mathrm{Var}\!\left(v_t(S\cup\{z\})-v_t(S)\mid S\cap \mathcal D_{\mathrm{irr}}\neq \varnothing\right)=0
\]
in the sense that this conditional branch is never visited. Equivalently, the entire contribution to variance arising from coalitions that include at least one irrelevant point is removed by restricting sampling to $S\subseteq \mathcal N(t)$. The corollary follows.
\end{proof}

%% file: sections_b_experiment_main.tex
\section{Detailed Experimental Setup}
\label{apx:exp}
\input{sections_b_experiment_i_setup}

%% file: sections_b_experiment_i_setup.tex
\subsection{Evaluated Models with Diverse Locality}

To evaluate the generality of model-induced support sets (Section~\ref{sec:model}), we instantiate locality across four representative model classes spanning distinct architectural mechanisms. For each model, we explicitly construct the support set $\mathcal{N}(t)$ according to its computational pathway, enabling direct measurement of support size, overlap, and distinct subset complexity.

\begin{itemize}
    \item \textbf{Weighted $K$-Nearest Neighbors (WKNN)}~\cite{dudani1976distance}.  Locality arises from geometric proximity in feature space. The support set contains the $2K$ nearest neighbors of $t$ under inverse-distance weighting with $K=5$.\footnote{Under the threshold formulation, this yields exact locality and serves as a reference case where support reduction is theoretically precise.}
    \item \textbf{Decision Tree}~\cite{breiman1984classification}.  
    Locality arises from rule-based partitioning. The support set consists of training instances that pass through the same parent node as the leaf reached by $t$, reflecting structural sparsity induced by decision paths. The support size is controlled through the \texttt{min\_samples\_split} and \texttt{min\_samples\_leaf} parameters in scikit-learn.
    \item \textbf{RBF Kernel SVM}~\cite{cortes1995support}.  
    Locality arises from kernel-induced decay of influence. The support set consists of training instances whose kernel value with $t$ exceeds a threshold, i.e., $\mathcal N(t) = \{ z \in \mathcal D : K(x_z, x_t) \ge 0.5 \}$.
    This setting captures approximate locality governed by bandwidth.
    \item \textbf{Graph Neural Networks (GNNs)}~\cite{kipf2017gcn}. 
    Locality arises from message passing over graph topology. The support set consists of nodes within the two-hop ego-network, and we train a two-layer GCN. This depth follows the standard configuration for node classification on homophilic citation graphs and is motivated by the oversmoothing literature~\cite{li2018deeper,oono2019graph}, which shows that deeper GCNs collapse representations and degrade performance in this regime.To match standard batching dynamics, we compute the local game over each training node’s support set and evaluate the corresponding test nodes within it.
\end{itemize}

These models span geometric proximity, dual sparsity, rule-based partitioning, and graph propagation. This diversity allows us to assess whether the support-set abstraction consistently induces measurable reductions in coalition space and retraining complexity beyond nearest-neighbor settings~\cite{jia2019towards,wang2024efficient}.

\subsection{Datasets}
\input{sections_6_experiment_table_tab_datasets}
We evaluate across diverse data modalities and model families, including image, tabular, and graph data, to systematically examine how different locality mechanisms manifest under representative structural conditions. For each setting, we pair the model family with a dataset whose structure aligns with its locality mechanism, ensuring that the support-set abstraction is tested under representative use cases. Dataset statistics are summarized in Table~\ref{tab:datasets}.

For WKNN, we use \textbf{MNIST}~\cite{lecun1998gradient}, randomly sampling 1{,}000 training and 1{,}000 test images and extracting 1{,}024-dimensional features via a pretrained CNN encoder~\cite{jia2019efficient}. The high-dimensional feature space provides a challenging setting for distance-based locality, as neighborhood structure becomes increasingly sensitive to the metric and dimensionality.

For Decision Tree, we use \textbf{Iris}~\cite{fisher1936iris}, a classical multiclass benchmark with 150 instances and four continuous attributes across three balanced classes. A 70:30 stratified split yields 105 training and 45 test instances. The compact feature space and clear class boundaries make it well suited for evaluating rule-based partitioning locality, where support sets are determined by decision paths.

For RBF kernel SVM, we adopt \textbf{Breast Cancer}~\cite{street1993nuclear}, comprising 569 samples with 30 real-valued diagnostic features extracted from digitized cell nuclei images. Under the same 70:30 split (398 training, 171 test), the moderate dimensionality and smooth class boundaries provide a natural testbed for kernel-induced locality, where influence decays with feature-space distance.

For GNN, we evaluate on \textbf{Cora}~\cite{mccallum2000automating}, a citation network with 2{,}708 nodes, 5{,}429 edges, and 1{,}433-dimensional bag-of-words features across seven classes. We designate 1{,}000 nodes as the test set and train on the remaining 1{,}708 nodes. The graph topology induces structural locality through message passing, allowing us to evaluate the framework in a non-Euclidean domain where neighborhoods are defined by graph distance rather than feature-space proximity.

\subsection{Baselines}
\label{app:baselines}
We implement four baselines within the standard Monte Carlo framework for Shapley value estimation.
\begin{itemize}
    \item \textbf{Global-MC~\cite{jia2019towards}} is the standard Monte Carlo Shapley estimator that operates over the full training set.
    
    \item \textbf{Local-MC} restricts the permutation space to the support set $\mathcal{N}(t)$ of each test point $t$, thereby reducing per-round retraining cost. Within this local region, the procedure is identical to Global-MC. The support set is defined in a model-specific manner following Section~\ref{sec:preliminary}. All other convergence and sampling parameters are shared with Global-MC.
    
    \item \textbf{TMC-S~\cite{ghorbani2019data}} augments Global-MC with early stopping to reduce the effective permutation length. Marginal contributions are computed sequentially within each permutation, and evaluation is truncated once the running prediction meets the stopping criterion under a performance tolerance (i.e., accuracy change falls below $0.01$ or the predicted class remains unchanged for five consecutive additions). All other convergence and sampling parameters are shared with Global-MC.
    
    \item \textbf{Comple-S~\cite{sun2024shapley}} estimates Shapley values via paired coalition evaluations, requiring two model evaluations per round. All other convergence and sampling parameters are shared with Global-MC.
\end{itemize}
All the above baselines operate without model-induced locality or cross-support reuse. They either evaluate permutations over the full training set or restrict computation locally without reorganizing around shared subsets. As a result, overlapping coalitions across test points are repeatedly retrained, leading to higher runtime.

\paragraph{In-Run Data Shapley.}\label{app:in-run-comparison}
We additionally include In-Run Data Shapley~\cite{wang2024data} as a non-retraining baseline that attributes through the gradient signal of a single SGD trajectory using first- (\emph{o1}) and second-order (\emph{o2}) Taylor approximations. Because it requires gradient-based training, the comparison is restricted to our GNN setting; attribution is computed along an SGD trajectory with batch size as $64$, lr~$=$~$0.01$, and~$200$ epochs (test accuracy $0.887$, matching the full-data baseline).

\begin{table}[h]
\small
\centering
\caption{LSMR-A vs.\ In-Run Data Shapley on GNN,Cora.}
\label{tab:inrun-comparison}
\begin{tabular}{lccccc}
\toprule
Method & $r$ & $\rho$ & Acc@5\% & Acc@10\% & Time \\
\midrule
\textbf{LSMR-A}        & \textbf{0.516} & \textbf{0.376} & \textbf{78.6} & \textbf{83.9} & 7{,}241\,s \\
In-Run o1~\cite{wang2024data} & 0.344 & 0.317 & 37.3 & 39.9 & \textbf{144\,s} \\
In-Run o2~\cite{wang2024data} & 0.347 & 0.319 & 38.0 & 39.7 & 260\,s \\
\bottomrule
\end{tabular}
\end{table}

The two methods occupy different points on the speed--fidelity trade-off. In-Run completes attribution in 144--260\,s---much faster than LSMR-A---but at the cost of substantially lower correlation with the Global Shapley estimate ($r=0.347$ vs.\
$0.516$) and markedly weaker downstream selection (Acc@5\% of $38.0$ vs.\ $78.6$;
Acc@10\% of $39.7$ vs.\ $83.9$). Moreover, doubling the selected fraction from
5\% to 10\% yields only a marginal gain for In-Run ($37.3\!\to\!39.9$ and
$38.0\!\to\!39.7$), leaving its selection curve nearly flat and well below that
of LSMR-A across the reported budgets. This near-flat, low-accuracy behavior
reflects trajectory-bound attribution: top-scored nodes earn their rank by
aligning with the gradients of the full-data trajectory rather than by forming
a broadly informative standalone training subset, so accumulating more of them
adds little marginal predictive value. 
\subsection{Ablation of Efficiency Mechanisms}
\label{app:ablation}

To quantify how much each of LSMR-A's three efficiency mechanisms contributes to the overall runtime reduction, we run a four-configuration ablation in which the mechanisms are activated incrementally: \textbf{Global-MC} (the standard MC baseline), \textbf{$+$Locality} (permutations restricted to $\mathcal{N}(t)$), \textbf{$+$Subset-Centric} (intra-support reuse), and \textbf{LSMR-A} (full pipeline with pivot-based inter-support reuse). All four configurations share identical hyperparameters.

\begin{table}[h]
\centering
\small
\caption{Ablation of LSMR-A's three efficiency mechanisms. Each column adds one mechanism to the previous.}
\label{tab:ablation}
\resizebox{0.95\linewidth}{!}{
\begin{tabular}{ll rrrr}
\toprule
Model & Metric & Global-MC & $+$Locality & $+$Subset-Centric & LSMR-A \\
\midrule
\multirow{2}{*}{DT}
  & Time (s) & 1{,}446    & 243   & 152   & \textbf{111}   \\
  & \# Train & 2.8M       & 0.5M  & 0.3M  & \textbf{0.2M}  \\
\midrule
\multirow{2}{*}{GNN}
  & Time (s) & 200{,}860  & 45{,}378 & 16{,}075 & \textbf{7{,}241} \\
  & \# Train & 28.2M      & 12.0M    & 3.8M     & \textbf{1.7M}    \\
\bottomrule
\end{tabular}
}
\end{table}

Each mechanism produces a multiplicative compression by addressing a distinct source of redundancy:
\begin{itemize}[leftmargin=*, itemsep=2pt]
    \item \textbf{Locality} ($5.95\times$ on DT, $4.43\times$ on GNN) is the baseline localized strategy of Section~\ref{sec:exact}: restricting each permutation to $\mathcal{N}(t)$ reduces both the number of retrainings per sample (permutation length $|\mathcal{N}(t)|$ instead of $|\mathcal{D}|$) and the cost of each retraining (training subset bounded by $|\mathcal{N}(t)|$). The compression is therefore largest when $|\mathcal{N}(t)| \ll |\mathcal{D}|$, which holds across all four model families in our experiments.

    \item \textbf{Subset-Centric reformulation} ($1.60\times$ on DT, $2.82\times$ on GNN) realizes Lemma~\ref{lem:local-closedform} in Section~\ref{sec:subset-centric}: a single utility evaluation $v_t(S)$ is algebraically distributed to all players in $\mathcal{N}(t)$ via the closed-form weights, eliminating the player-by-player loop that the baseline formulation requires. The gain scales with $|\mathcal{N}(t)|$ because more players share each evaluation, which is why the compression is larger on GNN than on DT.

    \item \textbf{Pivot-based scheduling} ($1.37\times$ on DT, $2.22\times$ on GNN) implements the canonical-evaluator rule introduced in Section~\ref{sec:lsmr}: each distinct subset is trained exactly once across all test points whose supports contain it, with all other occurrences reusing the result. Two factors jointly determine the net gain. (i)~The average pairwise support overlap controls how many reuse opportunities exist. (ii)~The pivot mechanism itself adds bookkeeping cost (graph intersections, pivot lookups) that partially offsets the savings; this overhead matters less when the per-retraining cost is high, as on GNN, and matters more relative to the savings when each retraining is cheap, as on DT. 
\end{itemize}

Composed multiplicatively, the three mechanisms yield a total $13\times$ reduction on DT and $28\times$ on GNN.

%% file: sections_6_experiment_table_tab_datasets.tex
\begin{table}[t]
    \centering
    \caption{Dataset statistics.}
    \label{tab:datasets}
    \small
    \begin{tabular}{lccccc}
        \toprule
        \textbf{Dataset} & $|\mathcal{D}|$ & $|\mathcal{T}|$  & \textbf{Features} & \textbf{Classes} & \textbf{Domain} \\
        \midrule
        Iris          & 105 & 45 & 4 & 3 & Tabular \\
        Breast Cancer & 398 & 171 & 30 & 2 & Tabular \\
        MNIST         & 1{,}000 & 1{,}000 & 1024 & 10 & Image \\
        Cora          & 1{,}708 & 1{,}000 & 1{,}433 & 7 & Graph \\
        \bottomrule
    \end{tabular}
\end{table}